# A Theoretical Computer Science Perspective on Consciousness[1]

Manuel Blum and Lenore Blum[2]

## ABSTRACT

The quest to understand consciousness, once the purview of philosophers and theologians, is now actively pursued by scientists of many stripes. This paper studies consciousness from the perspective of theoretical computer science. It formalizes the Global Workspace Theory (**GWT**) originated by cognitive neuroscientist Bernard Baars and further developed by him, Stanislas Dehaene, and others. Our major contribution lies in the precise formal definition of a Conscious Turing Machine (**CTM**), also called a Conscious AI. We define the **CTM** in the spirit of Alan Turing's simple yet powerful definition of a computer, the Turing Machine (**TM**). We are not looking for a complex model of the brain nor of cognition but for a simple model of (the admittedly complex concept of) consciousness.

After formally defining **CTM**, we give a formal definition of consciousness in **CTM**. We later suggest why the **CTM** has the *feeling* of consciousness. The reasonableness of the definitions and explanations can be judged by how well they agree with commonly accepted intuitive concepts of human consciousness, the range of related concepts that the model explains easily and naturally, and the extent of its agreement with scientific evidence.

## INTRODUCTION

Thanks to major advances in cognitive neuroscience, science is on the brink of understanding how the brain achieves consciousness. In 1988, cognitive neuroscientist Bernard Baars proposed a Global Workspace Theory (**GWT**) of the brain, sketched its architecture, and outlined its implications for understanding consciousness. See (Baars B. J., 1988) and (Baars B. J., 2019). That, together with the invention of fMRI in 1990, and the seminal investigations by Francis Crick and Christof Koch (Crick & Koch, 1990) into the neural correlates of consciousness, helped shake off the taboo on the scientific study of consciousness. As a consequence, the quest to understand consciousness is now actively pursued by scientists of many stripes.[3]

---

[2] The work of Manuel and Lenore Blum was supported in part by CMU, in part by a sabbatical year from CMU at the Simon's Institute for the Theory of Computing, and in part by a gift from UniDT. Avrim Blum will be an author on the expanded version of this paper (Blum, Blum, & Blum, Towards a Conscious AI: A Computer Architecture Inspired by Cognitive Neuroscience, In preparation) which will contain a more in-depth description of the *Sleeping Experts' Algorithm* and how it is used in this context. Email addresses: mblum@cs.cmu.edu, lblum@cs.cmu.edu, and avrim@ttic.edu.

[3] There are various approaches to the study of consciousness. In addition to *neural correlates* (Dehaene & Changeux, 2011), these approaches include the *psychological* (James, 1890) and (Freud S. , 1900); *philosophical* (Dennett D. C., 1991) and (Chalmers, 1996); information theoretic *measures of consciousness* (Tononi, 2004) and (Tononi & Koch, 2015); and *structural* (Baddeley & Hitch, 1974). Our approach to consciousness is *architectural*. It is informed by and close in spirit to (Baars B. J., 1997) and employed by (Dehaene S. , 2014) to study neural correlates.

We study consciousness from the perspective of Theoretical Computer Science (**TCS**), a branch of mathematics concerned with understanding the underlying principles of computation and complexity.[4] **TCS** is our principal tool in defining the Conscious Turing Machine (**CTM**) as a formalization of neuroscientist Bernard Baars' Theater of Consciousness. The **CTM** is proposed for the express purpose of understanding Baars' Theater model and for providing a **TCS** framework to understand consciousness. In settling on this model, we look for simplicity not complexity, for a simple mathematical model sufficient to explain consciousness, not a complex model of the brain or cognition. Our approach, in the spirit of mathematics and theoretical computer science, proposes formal definitions to fix informal notions and deduce consequences.

An important major goal is to determine if the **CTM** can *experience* feelings not just *simulate* them. We investigate in particular the feelings of pain and pleasure and suggest ways that those feelings might be generated (Chapter **4**). We argue that even a complete knowledge of the brain's circuitry - including the neural correlates of consciousness - cannot explain what enables the brain to generate a conscious experience such as pain. We propose an explanation that works as well for robots having brains of silicon and gold as for animals having brains of flesh and blood. Our thesis is that in **CTM**, an explanation lies with the ***architecture*** of the system, its ***basic processors***; its **expressive inner language** that we call **Brainish**; and its ***dynamics*** (prediction, competition, feedback and learning); that make it conscious (Chapter **3**).

In his Global Workspace Theory (**GWT**), (Baars B. J., 1997) describes conscious awareness through a theater analogy as the activity of actors in a play performing on the stage of Working Memory, their performance under observation by a huge audience of unconscious processors.

Unlike **GWT**, the **CTM** will have just one and the same actor always on stage. That actor can ask or answer questions, make or respond to requests, or communicate information. Through a well-defined competition, an audience member with a response to a question or request, or with a question, request, comment, or information of its own, can send that actor the script she is to deliver.

Here is Baars' sketch of his model:

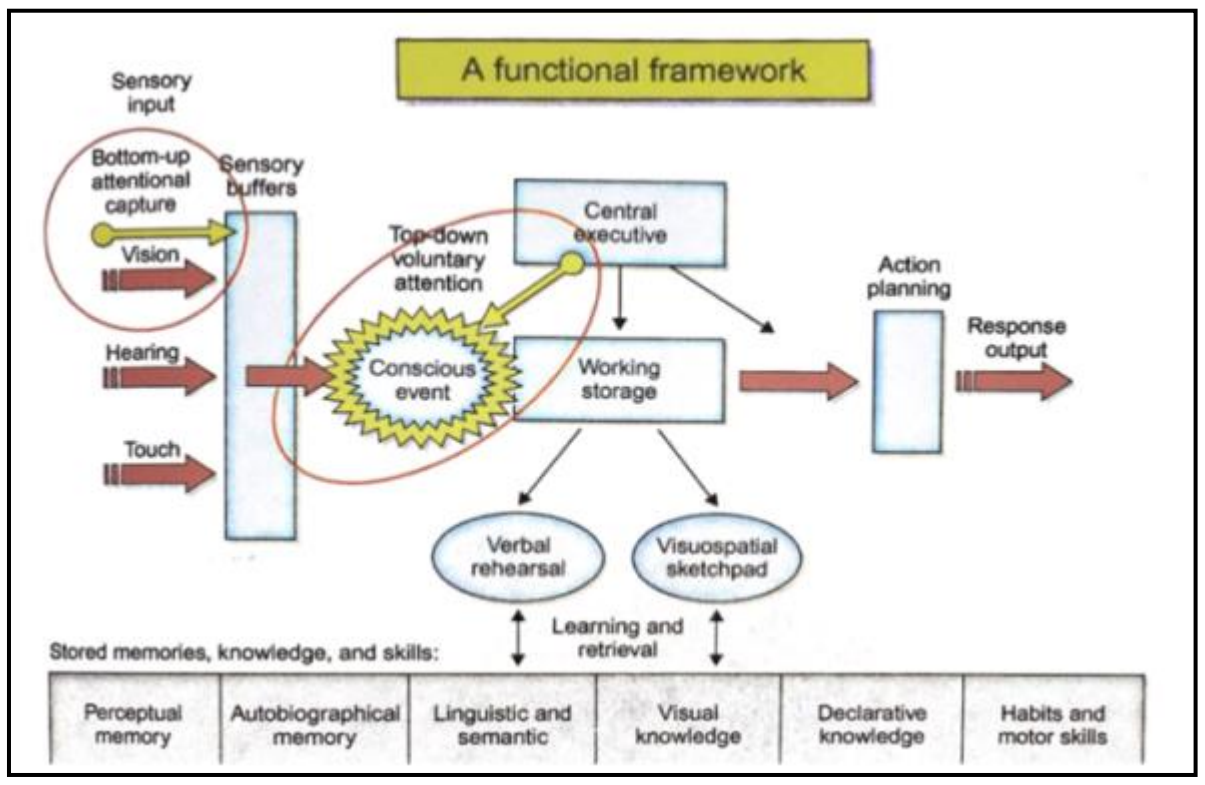

**Figure 1 Sketch of Bernard Baars' Global Workspace Model (adapted from** *(Baars & Gage, 2010)* **).**

---

The architectural approach to the study of consciousness was inspired by the architectural models of cognition, developed largely at Carnegie Mellon by Herb Simon's ***Sciences of the Artificial*** (Simon, 1969), Raj Reddy's **Blackboard Model** (Reddy, 1976), Allen Newell's ***Unified Theories of Cognition*** (Newell, 1990) and John Anderson's **ACT-R** (Anderson, 1996). **LIDA** (Baars & Franklin, 2009) is an important more recent architectural model of cognition.

[4] As a result of Gödel and Turing's groundbreaking works in the 1930's, many logicians became interested in the dichotomy between solvable and unsolvable problems (Hilbert's tenth problem). Starting in the 1960's and 1970's, theoretical computer scientists began pointing out that even amongst the solvable, there appeared to be a dichotomy between problems that are feasibly solvable (like maximum matching) and those that appear not to be (like SAT). The subsequent models and abstract theories of **TCS** led to remarkable insights into the mathematical distinction between efficiently and not efficiently solvable problems, an understanding of pseudo-randomness, applications to secure communication, machine learning, and much more. ( (Sipser, 2013) is a great introduction to **TCS**.)

A corresponding sketch of our **Conscious Turing Machine** (**CTM)** is in Section **1.7** (**The Conscious TM in Toto**) where we also discuss modifications and simplifications we have made to Baars' model. In the **CTM**, the stage is represented by a single Short Term Memory (**STM**), and the audience members sitting in the dark are represented by processors that make up its Long Term Memory (**LTM**). **LTM** processors, each with their own specialty, compete to get their questions, answers, and information on the stage.

The Turing Machine **(TM)**, being an extremely simple formal model of computation, is a fundamental first step in the mathematical understanding of computation.[5,6] In that spirit, **CTM**, being a formalization of cognitive neuroscientist Bernard Baars' theater model of consciousness, is a first step toward the theoretical computer science understanding of consciousness.

This formalization of **CTM** (Chapter **1**) includes the precise definition of a ***chunk***, a precise description of the ***competition*** that decides which (processor's) chunk will gain access to **STM**, and a precise definition of ***conscious awareness*** in the **CTM**. ***Feedback*** enables **LTM** processors to ***learn*** from their mistakes and successes; ***links*** that *emerge* in the life of the **CTM** enable conscious processing to become unconscious.

We begin by defining the **deterministic CTM** (Chapter **1**), which we use as a stepping stone to define the **probabilistic CTM** (Chapter **2**). For many reasons, (one given by Figure **3**/Note 4), we view the **probabilistic CTM**, not the **deterministic CTM**, as the simpler better model of consciousness. After Chapter **1**, when we refer to **CTM**, we mean the probabilistic variant unless we say otherwise.

The reasonableness of our formalization lies in the breadth of concepts that the **CTM** explains easily and naturally. That breadth includes some understanding for the *feeling of consciousness* (Chapter **3**) and the *Hard Problem* of consciousness which we explore in the particular case of pain and pleasure (Chapter **4**). The understanding depends on the *Brainish* language (Section **1.1**) which **LTM** processors use to communicate with each other, the activity of select **LTM** processors, and their *predictive dynamics* (prediction, feedback and learning), not on *chemicals* like glutamate, serotonin, dopamine, and so on.

Throughout this paper, we consider different scenarios that might arise in life, and investigate whether and how the model helps to explain the human experience of consciousness. The model does not explain everything. It is too simple for that. On the other hand, it explains a lot – and that without making any modifications to the basic **probabilistic CTM.**[7]

---

[5] The Turing Machine (**TM**), an example being the 23 state universal **TM** of (Minsky, 1967)**,** is a simple mechanism for exploring computability. The **CTM** is intended to serve a similar purpose for exploring consciousness.

[6] In his paper proposing the *Imitation Game*, aka the "Turing Test" (Turing, 1950), Alan Turing mentions consciousness, but only briefly, and basically skirts the topic: "In short then, I think that most of those who support the argument from consciousness could be persuaded to abandon it rather than be forced into the solipsist position. They will then probably be willing to accept our test. I do not wish to give the impression that I think there is no mystery about consciousness. There is, for instance, something of a paradox connected with any attempt to localize it. But I do not think these mysteries necessarily need to be solved before we can answer the question with which we are concerned in this paper."

[7] In our experience, no matter what property of consciousness one wishes to account for (see examples of properties in (Van Gulick, 2014)), no modifications of the basic **probabilistic CTM** model need be made. Instead, it suffices to invoke an appropriate **LTM** processor. Properties can be explained by the introduction of such processors, and these often work better than the otherwise obvious changes one might make to the model.

# TABLE OF CONTENTS

# 1 Formal Definition of the Conscious Turing Machine (CTM)

## 1.1 Preliminaries

Statements about the Conscious Turing Machine (**CTM**) are printed in black. Statements particular to humans or animals will generally be printed in burgundy. Burgundy-colored statements refer to features that a human or animal *would* have if it were correctly modeled by **CTM**.

**Time t** is discrete: **t = 0, 1, 2, ... T**. The **CTM** (its basic definitions given in Sections **1.2** and **1.3**) is **born** at time **t = 0** and has a fixed finite lifetime **T**. Time is maintained by a clock whose ticks are received by all components of **CTM** simultaneously.

**CTM**'s **world** is a high dimensional subset of $\mathbf{R^m(t) \times R^n(t)}$, where **R** is the reals, **m** and **n** are positive integer dimensions, and **t** is time. $\mathbf{R^m(t)}$ is **CTM**'s **outer world**, also called the **environment,** and $\mathbf{R^n (t)}$ is its **inner world.** **Input maps** transform the outer world to the inner world via **sensors**. These sensors may be for auditory, visual, tactile, thermal, gustatory, visceral, electromagnetic, or some other kind of information. **Output maps** transform the inner world to the outer world via **actuators** that act on the outer world.

**Brainish** is the **inner language** used by and between processors to communicate in its inner world. It includes coded representations of inputs and outputs all expressed with **multi-modal** Brainish words and phrases called **gists** (see Section **1.2.2.1**). Brainish is a much richer and more expressive language than **outer languages** such as English or Chinese for communicating in the outer world. Brainish is the language used to express **inner speech**, **inner vision**, and **inner sensations**. Its enormous expressive power can be appreciated by comparing wide-awake seeing with the seeing in **dreams**, as the dreams are manufactured completely by gists.

Gists can express and deal with images, sounds, tactile sensations, and thoughts - including unsymbolized thoughts[8]- and do this *better* than outer languages, which express only symbolized thinking (thoughts that can be communicated through the external environment). In humans, an important example of inner language is the collection of images, sounds, and actions that occur in dreams. A gist holds the essence of a scene in a nutshell. Having an expressive inner language is an important component of the *feeling* of consciousness (see Chapter **3**).

Besides Brainish, each processor in **CTM** has its own "inner" language for its own personal internal communication, dependent on its internal functioning. We say nothing more about each processor's own inner language here.

Although we use words like "small", "short", "succinct," and "fast" informally, they each have a technical meaning that will be specified when enough details have been given.

[8] Brainish can express both symbolized and unsymbolized language. (Hurlburt & Akhter, 2008) and (Vicente & Martínez-Manrique, 2016) give experimental evidence that human inner thought is always one (or two) of 1.speaking, 2.seeing, 3.feeling, 4.sensory awareness, and 5.unsymbolized thinking. (Hurlburt & Akhter, 2008) give the following example of an unsymbolized thought: "Abigail is wondering whether Julio (her friend who will be giving her a ride that afternoon) will be driving his car or his pickup truck."

## 1.2 Basic CTM Structure and Dynamics

**MAIN DEFINITION 1.2.1. CTM** is a **7**-tuple, **< STM, LTM, Down-Tree, Up-Tree, Links, Input, Output >**[9] where:

### 1.2.1 STM

is a **Short Term Memory** that at each and every time tick **t = 0, 1, 2, …, T** holds exactly one **chunk** (defined formally in Sections **1.3.1** and **1.3.2**). This single chunk becomes the entirety of **CTM**'s **conscious content** at time **t**. In humans, the storage capacity of short-term memory is roughly 7±2 chunks **(Miller, 1956)**, where a chunk can be a word, a phrase, a digit, and so on. A few chunks cycling through **STM** can simulate some aspects of an **STM** that holds several chunks.[10]

### 1.2.2 LTM

is a "large" collection of **N** (initially unlinked) **Long Term Memory** processors $p_1, p_2, \dots p_N$ whose workings are all **unconscious**. **Large** means that **N ≈ T**. Each processor, $p_i$, is a parallel random-access (programmable modifiable) machine with its own address, $\textbf{address}_{p_i}$ **= i**, and unbounded memory, $\textbf{memory}_{p_i}$. All **CTM** processors are in **LTM,** none in **STM,** none anywhere else, so "processor" will always mean **LTM** processor**.**

We assume that **LTM** has sufficiently many processors to guarantee that a fresh new unused processor is available or can be commandeered whenever a task requires one.

In a **CTM** with $N = 10^k$ processors, each processor has a **k**-digit address. We view the roughly $10^7$ cortical columns (k=7) in the human brain as constituting a substantial fraction of all its processors.

#### 1.2.2.1 LTM Processors Produce Chunks

At every clock tick **t = 0, 1, 2,…, T**, every **LTM** processor **p** produces a $\textbf{chunk}_{p,t,0}$, possibly a **NIL** chunk (Section **1.3.1**), which it places in the "competition" (defined formally in Section **1.3.2**) for **STM**. Chunks will be defined in Sections **1.3.1** and **1.3.2** as 6-tuples:

**chunk = < address**, **t**, **gist**, **weight**, **intensity**, **mood >.**

Each **chunk** contains the **address** of the processor that originated it, the time **t** it was generated, and a **gist** together with a real number **weight** (positive, negative or zero).

**Gists** are succinct compressed multi-modal thoughts. **Succinct** means that statements in Brainish are small enough to fit in a chunk, which in turn must be small enough to fit in any node of the **Up-Tree** (Section **1.2.4**). A gist can be an answer to a query, the (high level expandable) idea of a proof, an insight of some sort, a sketch of a beautiful sunset, a dream image, the pain of a torn ligament, and so on.

The **|weight|** of a gist is the processor's estimate of how important it is to get that chunk into **STM**. A **Sleeping Experts Algorithm** (Section **1.5.1**) that runs in each processor adjusts the weight-giving algorithms in that processor and the weights those algorithms give their gists. The Sleeping Experts Algorithm comes with the assurance that processors will eventually be neither too assertive nor too timid

---

[9] Coincidently, the classical **Turing Machine** is also defined as a **7**-tuple, $< Q, \Sigma, \Gamma, \delta, q_0, q_{accept}, q_{reject} >$, where **Q** is a finite set of **States**, **Σ** is the **Input** alphabet, **Γ** is the **Tape** alphabet, δ is the **Transition** function, $q_0$ is the **Start** state, $q_{accept}$ is the **Accept** state, and $q_{reject}$ is the **Reject** state.

[10] Cycling can happen via the **Up-Tree competition** (Section **1.3.2**) and the **Down-Tree broadcasts** (Section **1.2.3**). In this way, **CTM** can keep thoughts alive in **STM** continuously through many cycles by sending the thought from **processor → STM → processors → STM →** ….

in setting their weights (Blum, Hopcroft, & Kannan, 2015). The algorithm helps define the dynamics of feedback and learning.

The sign of the weight indicates whether the processor that created the chunk views the gist as positive/optimistic/cheerful or negative/pessimistic/depressing.

**Intensity** and **mood** will be defined in Sections **1.3.1** and **1.3.2**.

### 1.2.3 The Down-Tree

is a down-directed tree **/|** ↓ consisting of a single root in **STM** and **N** edges directed from that root to **N** leaves, each leaf being an input to one of the **N LTM** processors. At time **t**, $\mathbf{t \in \{0, 1, 2, \ldots, T\text{-}1\}}$, **STM broadcasts** its single chunk of content via the **Down-Tree** to all **N LTM** processors, causing them to receive that broadcast at time **t+1.**

**Conscious awareness** by **CTM** at time **t+1** is defined to be the reception by all **LTM** processors of the broadcast from **STM** at time **t** (see Section **1.6**). This ensures that processors responsible for the *sense* of conscious awareness (especially the model-of-the-world, inner speech, inner vision, and inner sensation processors) all receive the same content at the same time from **STM**.

While this is a formal definition of conscious awareness, it does not yet explain the *feeling* of conscious awareness (for that, see Chapter **3**).

### 1.2.4 The Up-Tree

is an up-directed binary tree[11]

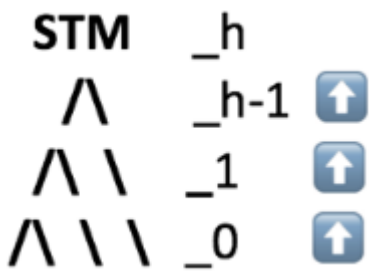

**Figure 2 Up-Tree.**

of height **h**. Its purpose is to run the **Up-Tree competition** (Section **1.3.2**) that determines which chunk gets into **STM**.

The **Up-Tree** has a single root in **STM** and **N** leaves, one leaf in each of the **N LTM** processors. Every directed path from a leaf to the root is *required* to have the same length **h**. We further require that $\mathbf{h \le 3(\log_2 N)}$. The **3** is arbitrary: any small integer **≥ 3** would do as well. (In Figure **2**, $\mathbf{h = 1.5(\log_2 N) = 3}$.)

Every node of the **Up-Tree** is at some level **s**, $\mathbf{0 \le s \le h}$. The leaf nodes are at level **0** and the root node is at level **h**. For each **s**, let $\mathbf{v_s}$ denote a node of the **Up-Tree** at level **s**. For **s > 0**, $\mathbf{v_s}$ contains a "small" "fast" parallel circuit with a "small" amount of storage that takes as input the chunks in its children and produces as output the chunk in $\mathbf{v_s}$ (details in Sections **1.3.2.1** and **1.3.2.2**).

At every time **t**, every processor puts a chunk in the **Up-Tree competition** (Section **1.3.2**) that begins at time **t** and ends at time **t+h** with a single winner, which is broadcast from **STM** to all **LTM** processors via the **Down-Tree**. **CTM** is constantly bubbling with the activity of chunks competing for **STM** and the winner of each competition being broadcast from **STM** to **LTM.** The time ordered chunks broadcast from **STM** to **LTM** form a **stream of consciousness** (Section **1.6**). As discussed in Chapter **3**, this stream is part of the subjective *feeling of consciousness*.

[11] In a binary tree, every non-leaf node has one or two children, no more no less.

### 1.2.5 Links

are bi-directional edges between processors. They form over time, and turn "conscious communication" between processors (via **STM**) into "unconscious communication" between them (through **links**).[12] For example, if processor **A** asks a question, **B** responds to it, and **A** acknowledges the response to be useful, and if this exchange occurs often enough, then a bidirectional **link** is generated between **A** and **B**.[13] **Links** transmit chunks and thus enable processors to influence each other directly, without going through **STM**. The number of two-way **links** between processors **A** and **B** at time **t** is proportional to the number of chunks from **A** broadcast by **STM** and acknowledged by **B** to have been useful plus the number of chunks from **B** broadcast by **STM** and acknowledged by **A** to have been useful, from time **0** to time **t**.

We note that a processor's address in a chunk provides negligible information about the processor's function, though processors may glean (some) information about such function from chunks broadcast from **STM** or communicated through **links**.

### 1.2.6 Input maps

take (time-varying) environmental information acquired by **CTM**'s sensors, convert that (bit-coded) information into Brainish-coded gists, then send those gists (encapsulated in chunks) to designated **LTM** processors.[14]

### 1.2.7 Output maps

convert Brainish-coded command gists from **LTM** processors (like those that generate instructions for a leg movement) into bit-coded commands for the intended actuators (the leg muscles). The (intended) action may or may not affect the environment, and even if it does affect it, may not affect it as intended or expected. The question whether or not an action has an effect and what that effect is, must be determined by **CTM** from feedback (observation of the environment) and learned experience (Section **1.5**).

[12] There are many examples in which **LTM** (the collection of unconscious processors) does some spectacularly heavy lifting, in part through unconscious communication. An example from Henri Poincaré is quoted in (Hadamard, 1945)**:** "As I was about to board a bus, the idea came to me, without anything in my former thoughts seeming to have paved the way for it, that the transformations I had used to define the Fuchsian functions were identical with those of non-Euclidean geometry."

[13] When a query is followed instantly by a response, the linking is that suggested by the Hebbian rule (Hebb, 1949): "Neurons that fire together wire together". In the **CTM**, however, the linking occurs even when the response comes some time after the query "Her name is Tina!"

[14] For simplicity, we assume that **sensors** are part of the **Input maps**, i.e. not separate entities. Similarly, we will assume **actuators** are part of the **Output maps**.

## 1.3 Important Details of the Up-Tree Competition

The **Up-Tree competition** (Section **1.3.2**) that starts at time **t** begins with each processor **p** putting a $\mathbf{chunk_{p,t,0}}$ on its leaf of the **Up-Tree** (Section **1.2.4**).[15] At each time $\mathbf{t > 0}$ and for every level **s**, $\mathbf{0 \le s < h}$, every chunk either moves up a level or disappears. These chunks, whether they "move up" or disappear, do so simultaneously in a single clock tick, meaning in the time interval **[t, t +1)**. The up or out decision depends on the **competition algorithm** and a chosen **competition function f** (Section **1.3.2.1**). The chunk at **s = h**, being the chunk in **STM**, is **broadcast** via the **Down Tree** to all **LTM** processors.

### 1.3.1 Chunks Submitted to the Competition

The chunk that processor **p** submits to the **Up-Tree competition** at time **t** is:

$$\mathbf{chunk_{p,t,0} = \langle\, address_p,\ t,\ gist_{p,t,0},\ weight_{p,t,0},\ intensity_{p,t,0},\ mood_{p,t,0}\, \rangle}, \text{ where}$$

a. $\mathbf{address_p}$ = address of the processor **p** that produces the chunk.

b. **t** is the time at which the chunk is submitted to the **competition**.

c. $\mathbf{gist_{p,t,0}}$ = the "small" amount of "information" in Brainish, also called the potential "thought", that **p** puts in $\mathbf{chunk_{p,t,0}}$ at time **t**. The potential thought becomes an actual thought one **tick** (a single unit of time) after it reaches **STM**.

   Sample gists include:

   - **Scene gist:** A rough sketch of a group with several well-dressed men and women talking together.
   - **Inner-speech gist:** "I don't know any of the people here."
   - **Feeling gists:** A sense of confusion.
     A desire to leave.
   - **Questions:** "Do I know any of these people?"
     "I know that woman, but… what's her name?"[16]

d. $\mathbf{weight_{p,t,0}}$ denotes the real number (positive, negative or zero) that processor **p** assigns to $\mathbf{gist_{p,t,0}}$ at time **t**.

---

[15] For simplicity, we have stipulated that all **LTM** processors submit chunks at all times to the competition for **STM**. In many cases $\mathbf{chunk_{p,t,0}} = \mathbf{chunk_{p,t-1,0}}$ ; in other cases, $\mathbf{chunk_{p,t,0} = \langle\, address_p, t, NIL, 0, 0, 0\, \rangle}$, the **NIL** chunk (defined in Section **1.3.1**). (**NIL** chunks have "negligible" effect on the "operation" of the **CTM**.) Unlike **CTM**, where all processors compete for **STM**, humans (and monkeys) have many processors that cannot compete. As (Milner, 2012) points out, for example, the ventral stream of vision is conscious (competes) while the dorsal stream is unconscious (does not compete):

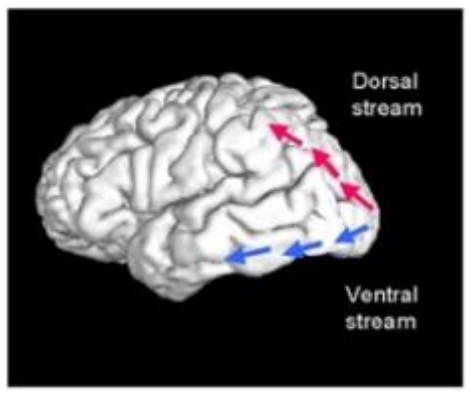

**Figure 2a Two Visual Systems - Goodale Lab - Western University (by permission of Mel Goodale).**

[16] An answer that might pop up after 1 minute: "Maybe the name begins with an S?"
An answer that might pop up a half hour later: "Now I remember, her name is Tina." If evidence of the answer's correctness is sufficiently strong (high enough |weight|), it provides feedback to all processors for correcting their answers and adjusting their weights (see Section **1.5.1**).

e. **intensity**$_{p,t,0}$ **= |weight**$_{p,t,0}$**|** is the importance that **p** assigns to its **gist**$_{p,t,0}$. It quantifies how much **p** "wants, needs, would like, or feels pressured" to broadcast **gist**$_{p,t,0}$ at time **t**.

Feedback is used (as we shall see in Section **1.5.1**) to drive the intensities that processors assign gists to reasonable values.

f. **mood**$_{p,t,0}$ = **weight**$_{p,t,0}$.

The empty or trivial gist is denoted by **NIL**. Any **chunk**$_{p,t,0}$ that contains the **NIL** gist at a weight of **0** is a **NIL** chunk.

For **t = 0**, we define $\mathbf{chunk_{p,0,0} = \langle address_p, 0, NIL, 0, 0, 0\rangle}$.

We have now defined **chunk**$_{p,t,s}$ for all **t** and **s = 0**. In the next section, we define **chunk**$_{p,t,s}$ for all **t** and $0 < s \le h$.

### 1.3.2 The Up-Tree Competition and The Chunks That Move Up

For $0 < s \le h$, **t = 0**, and for each node **v**$_s$, set $\mathbf{chunk_{p,0,s} = \langle address_p, 0, NIL, 0, 0, 0\rangle}$, where **p** is the *descendant* processor of **v**$_s$ having smallest address.

For $0 < s \le h$, **t > 0**, and for each node **v**$_s$, the **Up-Tree competition** places in **v**$_s$ at **time t+s**, **chunk**$_{p,t,s}$, for a particular **p**:

The particular **p** will be determined by the **competition algorithm** and **competition function f** (Section **1.3.2.1**). Once **p** is determined, the chunk in **v**$_s$ at time **t+s** will have the form:

$$\mathbf{chunk_{p,t,s} = \langle address_p, t, gist_{p,t,s}, weight_{p,t,s}, intensity_{p,t,s}, mood_{p,t,s}\rangle} \text{ where}$$

**gist**$_{p,t,s}$ = **gist**$_{p,t,0}$, **weight**$_{p,t,s}$ = **weight**$_{p,t,0}$, but **intensity**$_{p,t,s}$ and **mood**$_{p,t,s}$ are the sums of intensity and mood respectively of the chunks in the children of **v**$_s$.

**WARNING.** Thus for **s > 0**, unlike for **s = 0**, **intensity**$_{p,t,s}$ **≠ |weight**$_{p,t,s}$**|** and **mood**$_{p,t,s}$ **≠ weight**$_{p,t,s}$ in general.

**DEFINITION 1.3.2.1.** Processor **p wins the competition** that began at time **t** if **chunk**$_{p,t,h}$ (a variant of **chunk**$_{p,t,0}$) is in **v**$_h$ (i.e. **STM**) at time **t+h**. We call **chunk**$_{p,t,h}$ the **winning chunk**. (We may also call **chunk**$_{p,t,0}$ the **winning chunk**. What is meant is clear.)

The winning chunk, **chunk**$_{p,t,h}$, is **broadcast** via the **Down-Tree** to all **LTM** processors and simultaneously disappears from **STM**, all in the same single step in the time interval **[t+h, t+h+1)**.

The **Up-Tree competition** takes **h** time-units, **1** time-unit (the time between successive clock ticks) for each of the **Up-Tree's h** levels. This is quick, but not as quick as the **Down-Tree broadcast**, which takes just **1** time-unit.

#### 1.3.2.1 The Deterministic Competition Algorithm and the Competition Function

The **competition algorithm** is implemented by a collection of **N-1** circuits, one such circuit located in each of the **N-1** non-leaf nodes **v** of the **Up-Tree**. The circuit in each such **v** runs a *local* competition that selects (deterministically or probabilistically) one of **v**'s two children (siblings) based on a comparison of the chunks they contain, then moves (a variant of) that chosen child's chunk into **v**. The chosen chunk is said to be the **winner of the local competition** at/for **v**.

In specifying this algorithm, we aim for the local competition in each node to be run by a fast tiny parallel circuit.

The **competition algorithm** at **v** uses a **competition function**, **f**, to decide which of the chunks in **v**'s two children wins the local competition at that node. **f** is not used for any other purpose. The function, **f**, takes

$$\mathbf{chunk = \langle address, t, gist, weight, intensity, mood\rangle} \rightarrow \textbf{nonnegative real number}$$

in such a way that every node in the **Up-Tree** can do its computation and upload the winning chunk's information in the time between successive clock ticks. An example of **f** is **f(chunk) = intensity + ½ mood**. In the deterministic

**CTM**, the chunk with the bigger **f-value,** i.e. bigger **f(chunk),** moves up, unless its sibling has the same **f-value**, in which case the sibling with the smaller address moves up.

The deterministic **competition algorithm** in more detail:

Consider an arbitrary $\mathbf{v_s}$ in the **CTM Up-Tree**, $\mathbf{0 < s \le h}$. If $\mathbf{v_s}$ has just one child, let $\mathbf{v_{s-1}(L)}$ be that child, let $\mathbf{chunk_{p(L),t,s-1}}$ be the chunk in that child, and set **p = p(L)** and $\mathbf{chunk_{p,t,s} = chunk_{p,t,s-1}}$. Otherwise, $\mathbf{v_s}$ has two children, nodes $\mathbf{v_{s-1}(L)}$ and $\mathbf{v_{s-1}(R)}$, containing $\mathbf{chunk_{p(L),t,s-1}}$ and $\mathbf{chunk_{p(R),t,s-1}}$ respectively. If $\mathbf{chunk_{p(L),t,s-1}}$ has the larger **f**-value or if it has the same **f**-value as $\mathbf{chunk_{p(R),t,s-1}}$ and the smaller **address**, then set **p = p(L)**; else set **p = p(R)**. Then $\mathbf{chunk_{p,t,s-1}}$ is the **winner of the local competition** at level **s.** Then, at time **t+s**, the node $\mathbf{v_s}$ will contain the **chunk**:

$$\mathbf{chunk_{p,t,s} = < address_p,\ t,\ gist_{p,t,s},\ weight_{p,t,s},\ intensity_{p,t,s},\ mood_{p,t,s} >}, \text{ where}$$

$\mathbf{gist_{p,t,s} = gist_{p,t,0}}$, $\mathbf{weight_{p,t,s} = weight_{p,t,0}}$, $\mathbf{intensity_{p,t,s} = (intensity_{p(L),t,s-1}) + (intensity_{p(R),t,s-1})}$, and $\mathbf{mood_{p,t,s}} = \mathbf{(mood_{p(L),t,s-1})} + \mathbf{(mood_{p(R),t,s-1})}$. (Here we assume that if $\mathbf{v_s}$ has just one child, then that child contains $\mathbf{chunk_{p(L),t,s-1}}$, and $\mathbf{chunk_{p(R),t,s-1}}$ is **NIL**, so $\mathbf{intensity_{p(R),t,s-1}} = \mathbf{mood_{p(R),t,s-1}} = \mathbf{0}$.)

**NOTE 1.** By a simple induction on **s**, each specific $\mathbf{v_s}$ contains a $\mathbf{chunk_{p,t,s}}$ with $\mathbf{intensity_{p,t,s} = \sum (intensity_{p',t,0})}$ and $\mathbf{mood_{p,t,s}} = \sum \mathbf{(mood_{p',t,0})}$ where the two sums run over all **LTM** processors **p'** in the subtree rooted at this specific $\mathbf{v_s}$.

Thus, the **winning chunk** of the competition that began at time **t** is

$$\mathbf{chunk_{p,t,h} = < address_p,\ t,\ gist_{p,t,0},\ weight_{p,t,0},\ \textstyle\sum_{\text{all N processors p'}} (intensity_{p',t,0}),\ \sum_{\text{all N processors p'}} (mood_{p',t,0}) >}.$$

**NOTE 2**. The numbers $\mathbf{intensity_{p,t,h}/N}$ and $\mathbf{mood_{p,t,h}/N}$ are the **average intensity** and **average mood** over all chunks put into the competition at **time t**.

**NOTE 3**. We note that in the **deterministic competition** described above, the **winner of the competition** that began at time **t** depends on the specific assignment (permutation) of processors to the leaves of the **Up-Tree**:

**Example**. In the **deterministic competition** on four processors in Figure **3** whose chunks have gists **a, b, c, d,** with weights **3, 3, 1, 4,** respectively, and assuming **a's** address is a number smaller than **b's**, the chunk with gist **a** wins the competition. If the $2^{nd}$ and $3^{rd}$ processors are transposed so that their gists are **a, c, b, d,** then **d** wins.

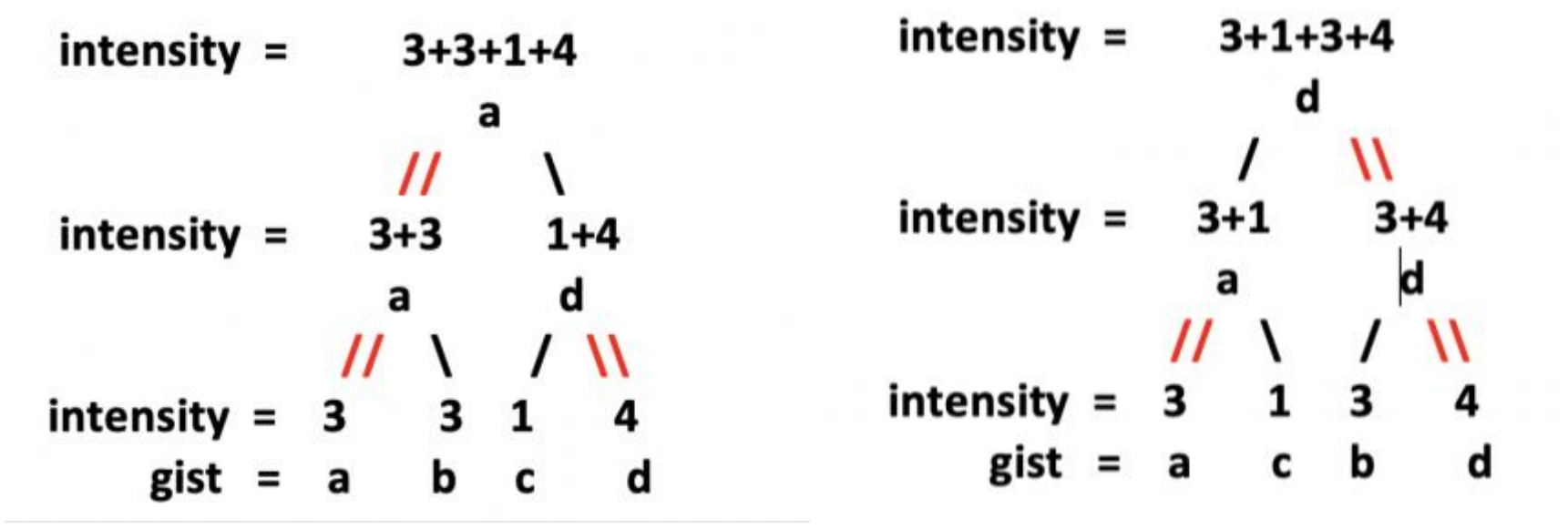

**Figure 3 Deterministic Competition with Competition Function f: chunk → intensity.**

**NOTE 4.** In the **probabilistic competition** to be introduced in Chapter **2** (Section **2.2**), local decisions are made with the aid of a **coin-flip neuron** (Section **2.1**). For "additive" competition functions **f** (Definition **2.2.1**), the winner of the competition will be *independent* of the assignment of processors to the leaves of the **Up-Tree**. In fact, something even better will hold true: the average fraction of time that a $\mathbf{chunk_{p,t,0}}$ gets to be in **STM** (in the form of $\mathbf{chunk_{p,t,h}}$) is given by $\mathbf{f(chunk_{p,t,0})} / \sum_{\text{all N LTM processors p'}} \mathbf{f(chunk_{p',t,0})}$ (see Theorem **2.2.1**).

### 1.3.2.2 The Competition Computation

For **t > 0** and **s > 0**, the **computation** to update the chunk at node $\mathbf{v_s}$ consists of doing all the following in **1 time-unit**:

**(i)** Evaluating which of the chunks associated with $\mathbf{v_s}$'s children, $\mathbf{v_{s-1}(L)}$ or $\mathbf{v_{s-1}(R)}$, has the greatest value of **f**, $\mathbf{f(v_{s-1}(L))}$ or $\mathbf{f(v_{s-1}(R))}$,[17] and if both have the same value, which has the smallest address, and choosing that one,

**(ii)** **putting** the **address, gist** and **weight** (but not the **intensity** and **mood**) of the chunk selected in (**i)** into the chunk at $\mathbf{v_s}$, and

**(iii)** **summing** the **intensities** and **moods** of the chunks associated with $\mathbf{v_s}$'s children, and setting those sums to be the **intensity** and **mood** respectively of the chunk at $\mathbf{v_s}$.[18]

**NOTE 5.** **(i)** above uses **f** to select a child, but once that child is selected, **(ii)** and **(iii)** make no further use of **f**. Every parent node $\mathbf{v_s}$ is a dedicated circuit that performs **(i)**, (**ii)**, and (**iii)**. These computations, all three of which must be completed in **1 time-unit**, put a bound on both *the size* of the **chunk** in a node and the *amount of computation* that can be performed in that node.[19]

**NOTE 6.** At the end of Chapter **2** (Section **2.3**) we discuss the **competition computation** to update the chunk at node $\mathbf{v_s}$ for the **probabilistic CTM**. Since the only difference in the models (**deterministic** vs. **probabilistic**) is how **local winners** in the **Up-Tree competition** are chosen (the **probabilistic CTM** utilizes a **coin-flip neuron**), the only difference in the computation will be in **(i)**.

The interested reader may take a detour here to visit **Chapter 2.**

## 1.4 More CTM Dynamics

Here we discuss some **CTM** dynamics that will play a role in **CTM**'s *feelings* of pain and pleasure (Chapter **4**).

### 1.4.1 The Interrupt Constant

Built into the **CTM** is a positive real number, the **Interrupt constant** $\boldsymbol{\iota}$. When a chunk gets to **STM**, it gets broadcast. If that chunk has **intensity** $\geq \boldsymbol{\iota}$, its reception causes all **LTM** processors to put their current work on a stack and to pay attention to the interrupt. So long as chunks passing through **STM** have **intensity** $\geq \boldsymbol{\iota}$, no processor can return to work *unless*, in its opinion, that work is potentially useful/ directly relevant/ tied to dealing with the interrupt. The Interruption of all processors coincides with the excruciating pain when a ligament is torn. In part, this is because the Model of the World processor (Chapter **3**) sees itself, actually its model of itself, in terrible pain. Section **4.1** discusses the role that the constant $\boldsymbol{\iota}$ plays in the "feeling" of pain.

A normal broadcast, unlike an interrupt, does not force any processor to put its work on a stack and pay full attention to it.

---

[17] This evaluation consists of two fast computations of **f** and a comparison of their values.

[18] The "linear operation" of summing was chosen because it is quick and yields many important features. For example, for **mood**, positive and negative valences cancel each other, while two negatives or two positives reinforce each other. The **CTM** is highly nonlinear, however, as a result of prediction, feedback, and learning (Section **1.5**), as well as the nonlinear intrinsic operations of the **LTM** processors themselves.

[19] The space required to store a chunk must be large enough to store a $\mathbf{log_2 N}$ bit (or $\mathbf{log_{10} N}$ digit) address, and to store a **gist** whose length is no greater than what is required to store approximately two lines of English or its equivalent in Brainish, very roughly $128 \times 128 = 2^{14}$ bits.

### 1.4.2 Increasing Weights

The **CTM** is built to look for ways to increase weights, whether positive or negative e.g. **+2** → **+3** and **-2** → **-1**. In the **CTM**, anything that increases weights is noted, learned and viewed as "pleasure". The baby learns that milk when hungry counters the negative weight of hunger. After that, whenever and for whatever reason the baby is in pain, it looks for the breast to reduce the pain. See (Leknes & Tracey, 2008) and (Harrison, et al., 2016). In Section **4.2** we argue that the dynamics of aiming to increase weights plays a role in the "feeling" of pleasure.

### 1.4.3 The High Level Story

We assume that at each time **t**, each processor **p** stores in its internal memory a tuple consisting of the $\mathbf{chunk}_{p,t,0}$ it submitted to the competition at time **t**, and all chunks it received at time **t**, whether by broadcast (as $\mathbf{chunk}_{p,t-h,h}$) from **STM**, from links, or from input maps. This history is necessary for the operation of the sleeping experts algorithm. In addition, it can contribute to a **high level story** of what **p** saw and did. Periodically, this stored information may be pruned so only "salient" chunks remain, the most "salient" being those that represent unexpected, bad (breaking a bone), or wonderful (birthday party) events. We do not specify details of this pruning process, if any, in the **CTM**.

## 1.5 Predictive Dynamics

In this section, we discuss **predictive dynamics**[20] **= prediction + feedback + learning** in the **CTM.**

- **Predictions** in **CTM** are made by each and every **LTM** processor. These are made both inside the processor's internal algorithms and through its connections to outside the processor – as it submits chunks to the competition for **STM**, to other processors via **links**, and to **actuators**.
- **Feedback** comes from chunks that are received through broadcasts from **STM**, through **links**, and from **sensors** of the outer world via **Input maps**.
- **Learning** and **Correcting** takes place within processors.

There is a continuous cycling of prediction, feedback and learning within **CTM**. In Chapter **3**, we argue that the "feeling" of consciousness in the **CTM** arises in part from this cycling.

Suppose you command your walk processor to walk you from home to work. With experience, that walk can be accomplished automatically, unconsciously, and your de facto prediction at the start of the walk can be a straight-forward walk with no interruptions. During the walk, attention can be paid elsewhere. Stumbling and skinning a knee, however, invalidates your prediction. It serves as feedback to the processor that made the prediction.[21] Next time you go out, you may avoid or pay special attention to that section of sidewalk.

[20] This is related to "predictive processing." See, in particular, (Lee & Mumford, 2003), (Friston, 2003), (Friston, 2005), (Cleeremans, 2014), (Clark, 2015), (Seth, 2015) and (Hohwy & Seth, 2020). Also see the earlier, "nets with circles", in (McCulloch & Pitts, 1943).

[21] Question: Does the walk processor call attention to the stumble (you expect to be upright when you plant your left foot in front of your right, but instead you find yourself falling), or does the pain processor call your attention? Put another way, do you notice you are stumbling before or after your knee gets skinned? In the late 1880's, William James asked a similar question (see (Ananthaswamy, 2015, p. 149)).

### 1.5.1 Sleeping Experts Algorithm

**Sleeping Experts Algorithms** are a class of learning algorithms employed by **LTM** processors.[22] Here we present one of the simplest versions of the **Sleeping Experts Algorithms**[23] for correcting errors in processors that generated faulty chunks.

Recall that as part of its **high level story** (Section **1.4.3**), every processor maintains a list of all chunks it has ever submitted to the competition. Each chunk is initially stored "unchecked". The following **Sleeping Experts Algorithm** checks off some of these chunks:

Fix **t > 0**. At whatever time **t' > t + h** a processor **p** learns (via broadcasts from **STM, links** or otherwise) that its submission to the competition at time **t** (whether or not that submission reached **STM**) was right or wrong, and provided that that submission has not yet been checked off, **p** does the following:

1. If what got to **STM** at time **t+h** was right, then **p** does nothing.
2. If what got to **STM** at time **t+h** was wrong, and
   - if **p** was right at time **t**, then
     - **p** promotes itself, i.e. increases its intensity giving power (say by multiplying it by **3/2**), and
     - **p** checks off this submission.
   - if **p** was wrong at time **t**, then
     - **p** corrects its error to the extent it can (e.g. her name was Tina, so it did not begin with S),
     - **p** then demotes itself, i.e. lowers its intensity-giving ability (say by multiplying it by **1/2**),[24, 25] and
     - **p** checks off this submission.

The way that processor **p** can tell that it was right or wrong at time **t**, and that what got into **STM** at time **t+h** was right or wrong, is from *feedback* **p** receives at some time **t' > t + h**, which would come not only from what **STM** broadcasts but also from what **p** receives from other processors via **links** and from the environment via **Input maps**. In general, every processor looks and judges for itself if it was in error.

Beliefs can be corrected and recorrected repeatedly. To see how this happens, consider the "i before e" rule for English spelling:

> i before e except after c, or when sounded like a as in neighbor or weigh, with exceptions such as *w**ei**rd*, *polic**ie**s*, *n**ei**ther*, *s**ei**ze*, nor *forf**ei**t*, ***ei**ther*, *caff**ei**ne*, *alb**ei**t, glac**ie**r, spec**ie**s*.

Now imagine there is one processor for the whole "i before e" rule, and separate processors for each word whose spelling must be remembered.

The first few times the processor $\mathbf{p}_{\text{caffeine}}$ for the correct spelling of "caffeine" sees that the misspelling of the word got into **STM** the algorithm raises the intensity giving power of $\mathbf{p}_{\text{caffeine}}$ until it overrides the "i before e" rule. Once its correction is high enough to override, **CTM** stops making mistakes on "caffeine" and so the algorithm stops raising $\mathbf{p}_{\text{caffeine}}$'s intensity giving power.

---

[22] These **Sleeping Experts Algorithms** are typically viewed as centralized procedures for adjusting |weights| on basic predictors (called "specialists" or "sleeping experts") that at any given time may make a prediction or abstain. In our setting, these **Sleeping Experts Algorithms** will be implemented in a distributed fashion, with each processor made responsible for correcting its own weights.

[23] More sophisticated **Sleeping Experts Algorithms** will be presented in an expanded version of this paper. See also, (Blum A. , 1995) (Blum A. , 1997), (Freund, Schapire, Singer, & Warmuth, 1999) , (Blum & Mansour, 2007) and (Blum, Hopcroft, & Kannan, 2015).

[24] The **3/2** and **1/2** are chosen so that an equal numbers of increases and decreases do not affect the average weight.

[25] If what got to **STM** was different from **p**'s gist, then **p** may choose to not demote itself or to demote itself less.

## 1.6 Definition of Consciousness in the CTM

Psychologists have defined consciousness as awareness of sensory stimulation - as opposed to merely being awake and receiving stimulation. "We are not conscious of everything we see and hear, nor of all of the information processing occurring in our own brains. We are aware of only a small subset of input and processing, which is woven together into a continuous and seamless narrative that we experience." (Novella, 2010)

Here we present our definitions of consciouness in the **CTM**, some of which have been stated earlier. Again, these are *formal definitions*; in Chapter **3** we discuss what generates the *feeling* of conscious awareness in **CTM**.

**DEFINITION 1.6.1.** At each time **t**, **t ≥ 0**, **STM** holds exactly one **chunk**, which is designated to be the entirety of **CTM**'s **conscious content** at time **t**. **Conscious awareness** in **CTM** of that chunk, which is **broadcast** from **STM** at time **t,** is defined to be its reception by all **LTM** processors at time **t+1**.[26, 27]

**One reason** to keep the number of chunks in **STM** small (exactly one in our model), and to keep the amount of information in any gist (and hence in any chunk) small (at most two lines of Brainish), is to ensure that all processors focus on (are consciously aware of) the same information in the broadcast from **STM**.[28] Equivalently, permitting just one chunk at a time into **STM** focuses the "feeling of consciousness" (see Chapter **3**) that occurs when all processors pay attention to the same tightly circumscribed content. A **second reason** is that while it might seem preposterous that the theater model could succeed with no central executive and just one actor on stage, these restrictions together with feedback and learning (Section **1.5**) underscore how the model succeeds even in this extreme case, supporting our **third reason**: to keep the model as simple as reasonably possible. Leslie Valiant (Valiant, 2013, pp. 127-128) views limited computational resources and constraints imposed by the need to learn as the primary reason for the small size of conscious information.

Since **CTM** becomes **consciously aware** of the **winning** $\mathbf{chunk}_{p,t,h}$ at time **t+h+1**, it follows that that there is a **delay** of **h+1** time units for **CTM** to become consciously aware of the winning **gist** that was submitted to the competition at time **t**, as well as the (**average**) **mood** of all gists that were submitted at time **t**. Neuroscience research demonstrates that there are time delays of **300msec** or more, between the time that decisions are made by unconscious processors and the time that humans first feel that they consciously made them (Libet B. (., 1985)**,** (Bode, et al., 2011) and (Guggisberg & Mottaz, 2013)**.**

At all times **t > 0**, the **CTM** is continuously active with chunks competing in the **Up-Tree** to get into **STM** and the chunk in **STM** being broadcast to all **LTM** processors. Thus **CTM** is continuously consciously aware of the changing content of **STM**.

**DEFINITION 1.6.2.** The **stream of consciousness** $(\mathbf{t_1}, \mathbf{t_2})$ is the sequence of chunks broadcast from **STM** to **LTM** in the time between $\mathbf{t_1}$ and $\mathbf{t_2}$. We describe this as the **stream of consciousness** without a time interval when that time interval is irrelevant.

---

[26] Note that while cognitive neuropsychology literature, e.g., (Graziano, Guterstam, Bio, & Wilterson, 2020), distinguishes between **conscious awareness** and **attention**, the **CTM** makes no such distinction.

[27] Blindsight provides a striking example of the difference between conscious and unconscious awareness (Striemer, Chapman, & Goodale, 2009). In blindsight, the **CTM** (or person) does not *consciously* see the outer world. Information from the **Input sensors** (eyes) goes directly to a subset of **LTM** processors related to vision but does not get up to **STM** (due to some malfunction, perhaps a break in the **Up-Tree**) and (therefore) *does not get* **broadcast**. For this reason, **CTM** (she) is *not* **consciously aware** that it (she) can see. However, information can be communicated between (unconscious) processors via **links**. For example, orders can be given by the **Walk Processor** to the leg actuator to take a walk that avoids obstacles. At a high level, this suggests how the blind-sighted person can have the surprising ability to avoid obstacles, despite that she believes herself to be blind.

[28] At the other extreme, consider an **STM** sufficiently large to contain a chunk from each of the **N** processors. That **CTM** clearly has problems focusing attention.

The constant **bubbling** of chunks competing to get up into **STM**, together with the continual broadcasting of successive winners down to **LTM**, produces the **stream of consciousness**. As in humans (James, 1890), this dynamic stream helps give **CTM** the "feeling" of consciousness including its richness and texture (see Chapter **3**).

The stream of consciousness is sustained by the constant **bubbling** of chunks as suggested by the (Gazzaniga, 2018) metaphor:

> "Each mental event is managed by brain modules [**CTM** processors] that possess the capacity to make us conscious of the results of their processing. The results [**CTM** chunks] bubble up from various modules like bubbles in a boiling pot of water. Bubble after bubble, each the end result of a module's or a group of modules' processing, pops up and bursts forth for a moment, only to be replaced by others in a constant dynamic motion. Those single bursts of processing parade one after another, seamlessly linked by time. … "

### 1.6.1 The Current Mood

Conscious awareness in **CTM** affects **CTM**'s mood.

**DEFINITION 1.6.1.1.** Let $\mathbf{t > h + 1}$. The **current mood** of **CTM** at time **t**, $\mathbf{mood_t}$, is defined to be the **mood** of the **chunk** that is broadcast from **STM** at time **t-1** and received by **LTM** processors at time **t** (so **CTM** is consciously aware of it at time **t**). Similarly, the **current intensity** of **CTM** at time **t**, $\mathbf{intensity_t}$, is defined to be the intensity of the **chunk** that is received by **LTM** processors from **STM** at time **t**.

**NOTE.** $\mathbf{mood_t} = \sum_{\text{all N LTM processors p}} \mathbf{mood_{p,t-1-h,0}}$ and $\mathbf{intensity_t} = \sum_{\text{all N LTM processors p}} \mathbf{intensity_{p,t-1-h,0}}$. So $\mathbf{mood_t/N}$ and $\mathbf{intensity_t/N}$ are the averages, respectively, of the moods and intensities of the chunks that were submitted to the competition at time **t-1-h**.

$\mathbf{Mood_t}$ is the measure of **CTM**'s **"**optimism/happiness" if positive, or "pessimism/sadness" if negative, at time **t.** (Kringelbach & Berridge, 2017) argue that in humans "emotion is always valenced—either pleasant or unpleasant—and dependent on the pleasure system". Similarly, $\mathbf{intensity_t}$ is the measure of **CTM**'s level of **"**energy/enthusiasm/confidence" at time **t**.

These measures are *formal definitions* of the stated feelings; they are not arguments that the **CTM** actually has those feelings. For that, see Chapters **3** and **4**.

**CTM**'s **current mood** globally affects the weights that processors assign to the gists that they submit to the **Up-Tree competition**. This happens in part because when a processor chooses the sign of a weight, if it is not clear what that sign should be, the sign is taken to be that of the $\mathbf{mood_t}$. Thus, a positive $\mathbf{mood_t}$ encourages positive thoughts, while a negative mood discourages them (leading to negative thoughts).

In addition, the **CTM** does the following: if the **current mood** is positive (negative), processors raise (drop) the weight they otherwise assign their "current" gist by a certain positive (negative) amount $\mathbf{\Delta \cdot w}$. Here $\mathbf{0 < \Delta < 1}$ is a constant of the **CTM**; **w** is the weight that **CTM** would otherwise assign the gist.

## 1.7 The Conscious TM in Toto

Chapter **1** focuses primarily on the basic architecture and dynamics of the **deterministic CTM**. The main components of **CTM's architecture** are **STM**, **LTM (processors)**, **Down-Tree**, **Up-Tree**, **Links**, **Input** (**maps**), **Output** (**maps)** and **chunks**. The main **dynamics** include chunk production, competition for **STM,** broadcasts to **LTM**, link formation, input output mapping, and learning based on prediction and feedback.

Chapter **2** defines the **probabilistic CTM**. The only difference between the deterministic and probabilistic **CTM** is that the probabilistic model uses a **coin-flip neuron** at every non-leaf node of the **Up-Tree**. The **probabilistic CTM** has the nice property that, when the **competition function f** is **additive** (Definition **2.2.1** of Chapter **2**), all chunks submitted to the **Up-Tree competition** get a fraction of time in **STM**. That fraction at time **t** is $\{ \mathbf{f(chunk_{p,t,0})} / (\sum_{\text{all N LTM processors } p'} \mathbf{f(chunk_{p',t,0})}) \}$, which is **chunk**$_{p,t,0}$'s share of its **f**-value, $\mathbf{f(chunk_{p,t,0})}$. That $\mathbf{f(chunk_{p,t,0})}$ is a measure of the importance and correctness of $\mathbf{chunk_{p,t,0}}$ as determined by **f**.

We are interested in the explanatory power of the basic **CTM**, whether it be deterministic or probabilistic. To this end, we single out several key processors essential for the *feeling* of consciousness (Chapter **3**). We do not focus on developing the **LTM** processors, though they are the heavy lifters that provide much of the functionality of the **CTM**.

Here is a sketch of our formal model, the **CTM**:

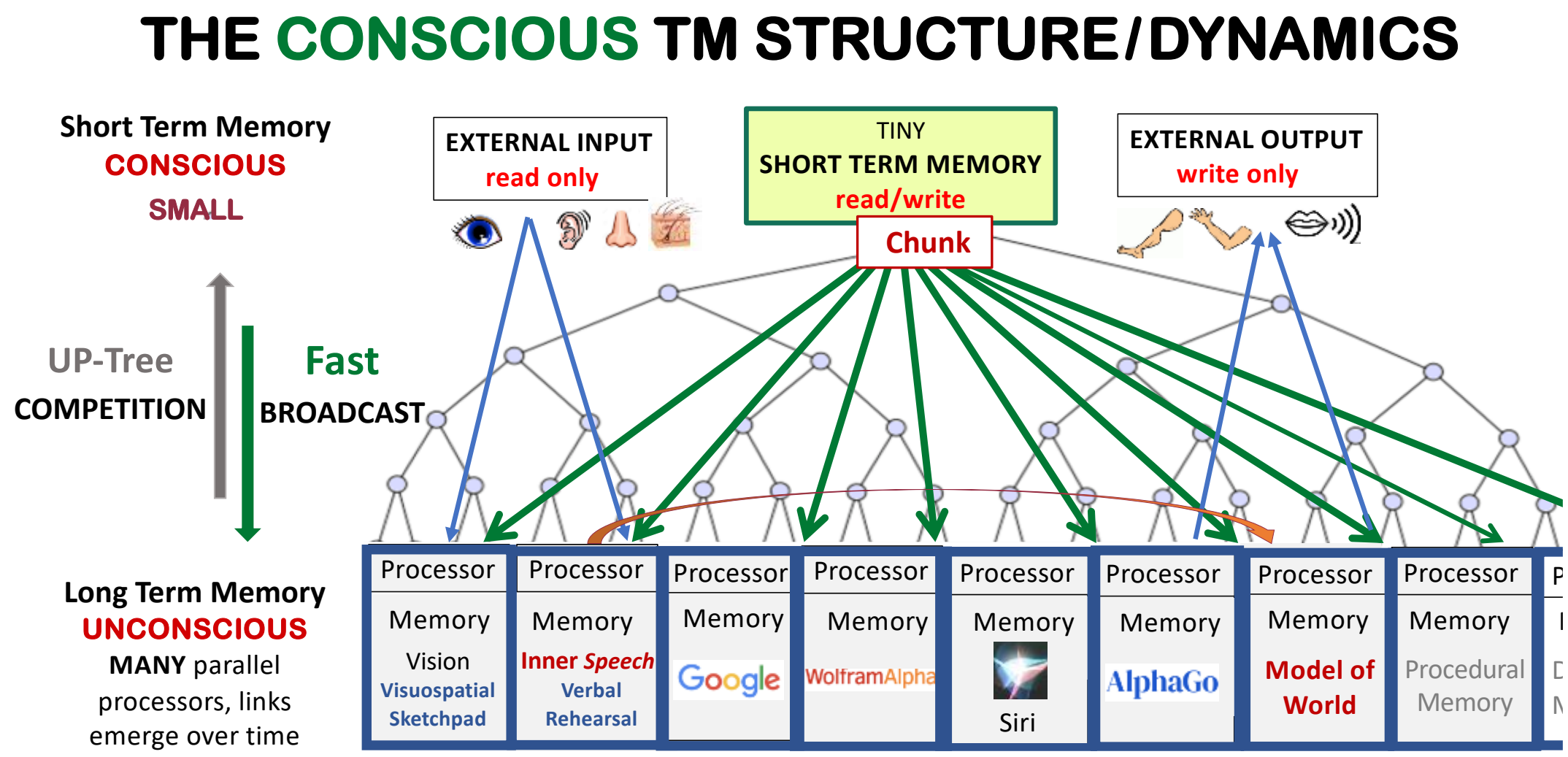

**Figure 4 CTM in Toto.**

In a comparison of **CTM** to **Baars' Theater Model** (Figure **1**), **Short Term Memory (STM)** is the stage. In **CTM**, there is always just one and the same actor on stage. At every step in time, that actor gets handed the winning chunk as a script for broadcast. The **Down-Tree** is the broadcast system, the **Up-Tree** is the competition process, and the **Long Term Memory (LTM)** is the audience of processors, each vying to get its chunk to the (actor on) stage.

Note that the **CTM** differs from **Baars' Theater Model** in several ways, including especially:

1. The **CTM** has no Central Executive aka stage manager.
2. In **CTM**, **Inputs** from the environment go directly to **LTM**, not to **STM**.
3. In **CTM**, **Outputs** to the environment are sent from **LTM**, not **STM**.
4. In **CTM**, unlike in humans, Working Memory = Short Term Memory.
5. In **CTM**, chunks compete in a *well-defined* competition to reach **STM**. That competition is vague in the Theater model.
6. In **CTM**, conscious awareness lies in the reception by all **LTM** processors of the content of **STM**. It is not an event that occurs between **Input** and **STM**.
7. Baddeley and Hitch prove the existence of a tiny **Speech Buffer** that they call the **Phonological Loop** for **Verbal Rehearsal** (Baddeley & Hitch, 1974), (Baddeley, 1986), (Baddeley, 2000) and (Baddeley, 2010).[29]

---

[29] If one needs to remember a phone number, one can repeat the number over and over until paper and pencil is found to write it down. This keeps the Speech Buffer full, making it difficult to use that buffer to find one's way around in the world. Finding paper and pencil is possible because the Visuospatial Sketchpad is available to help plan and implement this search.

They view the Phonological Loop as a component of **STM** not **LTM**. Baars places it between **STM** and **LTM**. To keep our model simple (rather than physiologically correct), **CTM** has no Phonological Loop (though it can recruit an **LTM** processor to do some of what the Phonological Loop does).

8. Baddeley and Hitch prove the existence of a **Visuospatial Sketchpad** as another component or slave of **STM**. To keep **CTM** simple (rather than physiologically correct), it has no Visuospatial Sketchpad (though it can recruit an **LTM** processor to do some of what the Visuospatial Sketchpad does).
9. As in the “Extended Mind Theory" of (Clark & Chalmers, 1998), **CTM** can have access to existing technology - Google, WolframAlpha, AlphaGo, NELL, and so on – in the form of **LTM** processors tasked to use these apps. This is one way to ensure that **CTM** has a huge collection of powerful processors at the start of its life, a collection that is also augmentable throughout life.

We note that key features of the formal **CTM** and its dynamics resonate with properties of consciousness that (Dennett D. C., 2018) outlines**:**

> [Neither] a Master Scheduler, nor a Boss Neuron, nor a Homunculus or Res Cogitans [govern the transitions of our conscious minds]. [What governs] must be a dynamical, somewhat competitive process of contents vying for fame, for cerebral celebrity ... or relative clout against the competition. What determines the winners? Something like micro-emotions, the strength of positive and negative valences that accompany and control the destiny of all contents, not just obviously emotionally salient events such as obsessive memories of suffering or embarrassment or lust, but the most esoteric and abstract theoretical reflections.

# 2 The Probabilistic CTM

The **CTM** defined in Chapter **1** is completely deterministic. Chunks that have low **f**-values, typically those for context and background, often do not get into **STM**. Humans, on the other hand, are generally conscious of context and background.

In the **probabilistic CTM** - but not the **deterministic CTM** - under reasonable conditions[30] on the relative importance of different chunks, a **chunk**$_{p,t,0}$ will win the competition for **STM** a fraction of time proportional to its importance (Theorem **2.2.1**).

The concepts and discussions regarding **CTM**, except as otherwise noted, apply to *both* deterministic and probabilistic models. The only difference between models is that the **probabilistic CTM** uses a **coin-flip neuron** (defined below) in every non-leaf node of the **Up-Tree**. Except for that, the **probabilistic CTM** is completely deterministic.

[30] The condition is that the competitive function **f** is additive (Definition **2.2.1**).

## 2.1 The Coin-Flip Neuron

**DEFINITION 2.1.1.** A **coin-flip neuron** is a device that takes as input an (ordered) pair **(a, b)** of non-negative real numbers (**a ≥ 0** and **b ≥ 0**), and in *one step* does the following:

if **a > 0** or **b > 0**, it outputs **a** with probability **a/(a+b)**, else **b**; else (**a + b = 0**) it outputs **a** with probability ½.

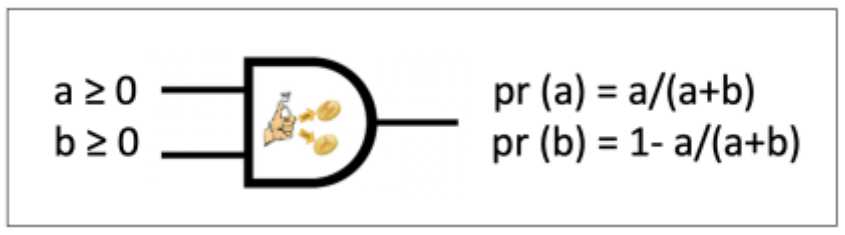

**Figure 5 A coin-flip neuron on input (a, b) with a + b > 0.**

In what follows, we **assume** that every node of the **Up-Tree**, at every level **s**, **0 < s ≤ h**, has a single **coin-flip neuron**.

## 2.2 The Probabilistic Up-Tree Competition

Let **f** be an **Up-Tree competition function** (Section **1.3.2.1**), i.e.,

**f:** chunk = < **address, t, gist, weight, intensity, mood** > → **nonnegative real number**

For the **probabilistic competition algorithm** (with **competition function f**):

The **address, t, gist,** and **weight** of the **chunk** to be associated with $v_s$, **0 < s ≤ h**, will be,

- with probability **f(chunk$_L$) / (f(chunk$_L$) + f(chunk$_R$))**, it will be the **address**, **t**, **gist**, and **weight** of the **chunk$_L$** associated with its left child **L**. In this case, we say that **chunk$_L$ moves up** and set **p** = **p(L)**.
- If **chunk$_L$** does not move up, then the **chunk$_R$** associated with its right child **R**, **chunk$_R$**, moves up and set **p** = **p(R)**.

The other parameters of the **chunk** associated with $v_s$ are defined, as for the deterministic competition (Section **1.3.2.1**), by

$\textbf{intensity}_{p,t,s}$ = ($\textbf{intensity}_{p(L),t,s-1}$) + ($\textbf{intensity}_{p(R),t,s-1}$) and $\textbf{mood}_{p,t,s}$ = ($\textbf{mood}_{p(L),t,s-1}$) + ($\textbf{mood}_{p(R),t,s-1}$).

**DEFINITION 2.2.1.** A competition function **f** is **additive** if for each **t**, **0 < t**, each **s**, **0 < s ≤ h**, each node $v_s$, and each $\textbf{chunk}_{p,t,s}$ in $v_s$, $\textbf{f}(\textbf{chunk}_{p,t,s}) = \textbf{f}(\textbf{chunk}_{p(L),t,s-1}) + \textbf{f}(\textbf{chunk}_{p(R),t,s-1})$.

In this case, it is convenient to define $+_f$ to be an operation on the **chunk$_L$** and **chunk$_R$** in sibling vertices that sets **chunk$_P$** in their parent vertex to be **chunk$_P$** = **chunk$_L$** $+_f$ **chunk$_R$**.

**Examples** of additive competition functions include**:**

$\textbf{f}(\textbf{chunk}_{p,t,s}) = \textbf{intensity}_{p,t,s}$, or more generally
$\textbf{f}(\textbf{chunk}_{p,t,s}) = \textbf{intensity}_{p,t,s} + \textbf{c} \cdot \textbf{mood}_{p,t,s}$ for any real **c**, **-1 < c < +1**,

but not the competition function $\textbf{f}(\textbf{chunk}_{p,t,s}) = |\textbf{mood}_{p,t,s}|$ as it is not additive, nor the function $\textbf{f}(\textbf{chunk}_{p,t,s}) = \textbf{mood}_{p,t,s}$ as it is not even a competition function.

We next show that the **probabilistic competition** with any additive competition function **f** gives every processor **p** a fraction of time in **STM** at time **t+h** that is proportional to the **f**-value of its chunk, $\textbf{f}(\textbf{chunk}_{p,t,0})$, at time **t**.[31] To see this through an example (Figure **6**), suppose **LTM** has **4** processors with chunks **a, b, c, d** having **f**-values **1, 3, 2, 4,** respectively. Then **b** will get a fraction of time in **STM** that is **3/(1+3+2+4)**:

[31] In this way, just for example, the environment (via inputs to processors) generally maintains a presence in **STM**.

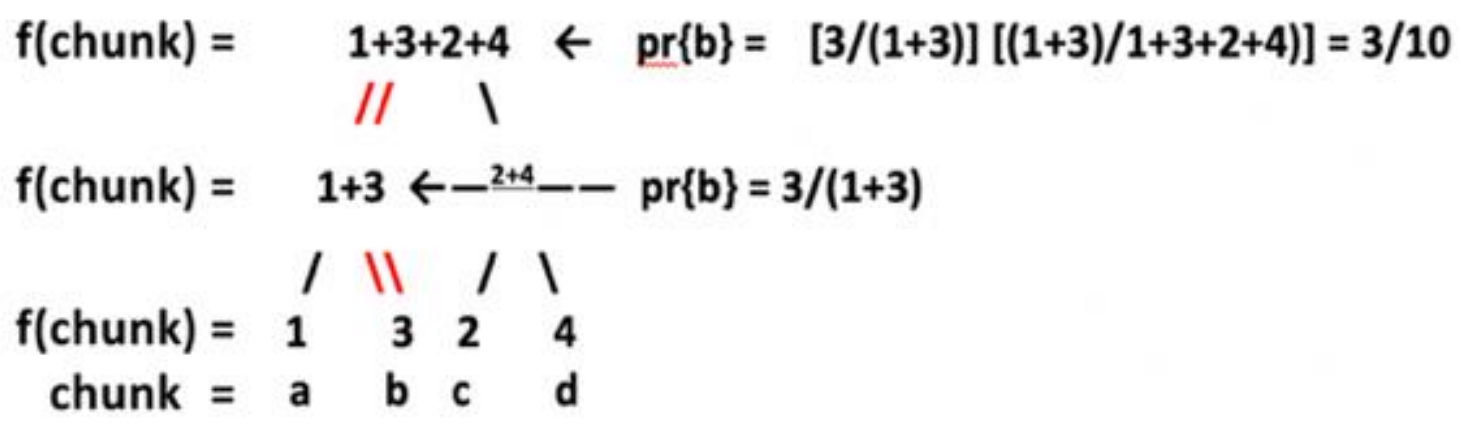

Figure 6 A Probabilistic Up-Tree.

**THEOREM 2.2.1** Let **f** be any additive competition function. Then for every processor **p** and time $\mathbf{t \geq 0}$, the probability that **p** wins the competition that began at time **t**, which by definition is the probability that $\mathbf{chunk_{p,t,h}}$ is in **STM** at time **t+h**, is $\mathbf{f(chunk_{p,t,0}) / \sum_{all\ N\ LTM\ processors\ p'} f(chunk_{p',t,0})}$. In symbols,

$$\mathbf{pr\ \{p \text{ wins the competition that began at time } t\}} =_{def} \mathbf{pr\ \{chunk_{p,t,h} \text{ is in STM at time } t+h\}}$$
$$= \mathbf{f(chunk_{p,t,0}) / \sum_{all\ N\ LTM\ processors\ p'} f(chunk_{p',t,0})}.$$

**PROOF:** Self-evident from the example. ■

**COROLLARY 2.2.1**. Let **f** be any **additive competition function**. Then for all **p** and **t**, the probability that $\mathbf{chunk_{p,t,0}}$ reaches **STM** at time **t+h** is independent of the location of **p** (or any other processor) on the leaves of the **Up-Tree**. Equivalently, the permutation chosen to assign processors to leaves of the **Up-Tree** has no effect on the sequence of broadcasts from **STM**.

While this convenient result holds for the competition function $\mathbf{f(chunk_{p,t,s}) = intensity_{p,t,s}}$ because this **f** is additive, it does *not* hold for the competition function $\mathbf{f(chunk_{p,t,s}) = |weight_{p,t,s}|}$, which puts a chunk having the largest **|weight|** at time **t** into **STM**.

Notice that for any additive competition function **f**, background chunks (i.e. chunks that get only a small fraction of time in **STM**) lurk constantly in consciousness, being (almost) completely out of consciousness only when **CTM** focuses intensely on something that needs full attention. This is one of several nice properties of the probabilistic **CTM** with additive competition function.

## 2.3 The Probabilistic Up-Tree Computation

In Section **1.3.2.2** we discussed the computation involved in updating the chunk at node $\mathbf{v_s}$ in the **deterministic Up-Tree competition**. Here we do the same for the **probabilistic Up-Tree competition**:

For **t > 0** and **s > 0**, the **computation** to update the chunk at node $\mathbf{v_s}$ consists of doing all the following in **1 time-unit**:

**(i)** Computing the **f**-value of the chunks associated with $\mathbf{v_s}$'s children, $\mathbf{f(v_{s-1}(L))}$ and $\mathbf{f(v_{s-1}(R))}$, and

a) if $\mathbf{(f(v_{s-1}(L)) + f(v_{s-1}(R))) \neq 0}$, choosing the left child $\mathbf{v_{s-1}(L)}$ of $\mathbf{v_s}$ with probability $\mathbf{v_{s-1}(L)/(f(v_{s-1}(L)) + f(v_{s-1}(R)))}$ else the right child $\mathbf{v_{s-1}(R)}$ of $\mathbf{v_s}$, or

b) if $\mathbf{(f(v_{s-1}(L)) + f(v_{s-1}(R))) = 0}$, choosing the left child of $\mathbf{v_s}$ with probability ½

**(ii)** **putting** the **address, gist** and **weight** of the chunk selected in **(i)** into the chunk at $\mathbf{v_s}$, and

**(iii)** **summing** the **intensities** and **moods** of the chunks associated with $\mathbf{v_s}$'s children, and setting those sums to be the **intensity** and **mood** respectively of the chunk at $\mathbf{v_s}$.

**NOTE.** The evaluation in **(i)** consists of two fast computations of **f**, a sum and division of their values, and a fast probabilistic selection. **(ii)** and **(iii)** are the same as the **deterministic competition**.

# 3 The Feeling of Conscious Awareness [32]

While **CTM** is consciously aware *by definition* of the broadcasted content of **STM**, this definition does not explain what generates the *feeling* of conscious awareness in **CTM**. This brings us to our big question: **Will CTM have the "feeling" that it is conscious?** While we believe that the answer is **YES**, we cannot prove anything mathematically without a definition of the "feeling of consciousness", which we do not have (yet).[33] Instead, we now present arguments for our belief that **CTM** has the "feeling" that it is conscious. In Chapter **4** we argue that **CTM** can have also the *feeling* of pain and pleasure.

Dreams play a special role in this argument.[34] That is because inputs and outputs are turned off in sleep, so what you see, hear, feel and do in a dream are creations – *fabrications* - of brain-generated gists. Dreams give a sense of the enormous power of Brainish to express sensations, actions, and feelings. The architecture of the **CTM** ensures that what you see in the environment has been coded by sensors of the environment into gists that go from those sensors to a few specialized **LTM** processors to **STM** and from there by broadcast to all **LTM** processors. As a consequence, dreams can be generated internally from memories using the same processing that enables the **Model-of-the-World processor** (defined below) to predict consequences of possible actions. In addition, dreams are conscious happenings that are capable of expressing emotions superbly.

We argue that the *feeling* of consciousness in **CTM** is a consequence principally of its extraordinarily expressive Brainish language, coupled with **CTM**'s *architecture,* certain *special processors*, and **CTM**'s *predictive dynamics* (prediction, feedback and learning).

More specifically:

1. The content of **STM** (i.e., the conscious content of **CTM**) is broadcast to all **LTM** processors, so all processors responsible for the feeling of consciousness know what's in **STM**.
2. Certain processors play a special role in generating the feeling of consciousness. Here we consider a few such processors that have their specialized algorithms built into them at birth:[35]
   - The **Model-of-the-World processor** is a collection of processors that construct models of **CTM**'s outer and inner worlds. We call each of these models a **model-of-the-world.** The **Model-of-the-World processor** tags various constituent parts of the model(s) as either **self**, **not-self**, or **unknown**. It also tags parts with additional descriptions such as the actions they can perform annotated in Brainish.[36]

---

[32] Discussions of consciousness often pit a*ccess* (or functional) consciousness against *phenomenological* (the subjective experience of) consciousness. (Block, 1995) sees these two types of consciousness as distinct. (Kriegel, 2006) argues that, while different, the former is a subcategory of the latter. We believe that subjective experience (e.g., *the feeling of what it is like to be me*) is possible in the **CTM**, and that the *explanatory gap* (Levine, 1983) can be filled. This viewpoint aligns closely with Baars (see (Kaufman, 2020) interview) and (Dennett D. C., 2016).

[33] We are exploring **IIT** (Tononi, 2004), among other theories, for a definition of the "feeling of consciousness". The **CTM** has a positive, **PHI**, **IIT**'s measure of consciousness, but does having consciousness, according to this measure, imply that it has the *feeling* of consciousness?

[34] Most people dream and most dreams have a large visual component. If you don't dream or don't remember your dreams, then our argument is not for you. If your dreams are auditory or tactile, you may be able to substitute that sense for the vision that we discuss.

[35] In the expanded version of this paper (Blum, Blum, & Blum, Towards a Conscious AI: A Computer Architecture Inspired by Cognitive Neuroscience, In preparation) we discuss these processors in more detail, as well as processors for **Sleep**, **Dreams, Meditation** and **Motivation**.

[36] For example, the notation attached to a leg representation might be "this leg can be moved with the power of thought; but this leg has no sensory feeling."

- The **Inner Speech processor** takes any speech (inner[37] or Brainish coded outer)[38] encoded in the gist broadcast by **STM** and maps it to the same location(s) that the **Input map** sends gists of outer speech.

  When the **CTM** begins speaking, speech from the **Inner Speech processor** passes through **STM**. After enough speech has passed through **STM**, that speech can go directly through **links**.

- **Inner Vision**[39] and **Inner Sensation processors** map whatever images/sensations (inner or Brainish coded outer) are broadcast from **STM** to whatever locations **input maps** send outer scenes/outer sensations. This enables **CTM** to see with its "mind's eye" what it sees with its actual sensory eye and to sense with its "mind's skin" what it senses with its actual sensory skin. (The **mind's eye** in the **model-of-the-world** "sees" whatever the **CTM** recalls from its visual memory. Similarly, the **mind's skin** in the **model-of-the-world** "senses" whatever **CTM** recalls from sensory memory.)

The **Inner Speech**, **Inner Vision,** and **Inner Sensation** processors are special purpose **decoders** that extract speech, vision, and sensation from the multi-modal gists that **STM** broadcasts.[40] They and the **Model-of-the-World processor** contribute to the *feeling* of consciousness as follows:

- The **Model-of-the-World processors** maintain models of the outer and inner worlds. They have several important jobs that give the **CTM** its *sense of self*, including:

  - Generating, recalling and maintaining (personal) maps of **CTM**'s worlds,
  - distinguishing **self** from **not-self** in those worlds,
  - helping to predict/correct actions of **self** and **not-self** in those worlds,
  - helping to plan actions in the environment (outer world),
  - labeling the objects in those worlds (in Brainish), and
  - labeling the **CTM** in its models of itself as "consciously aware", which it does when it detects itself thinking about its own consciousness.

  The **Model-of-the-World processor** can create and stitch together a sequence of chunks to produce an "inner movie", which sends images, smells and sounds to the appropriate (model-of-the-world) sensory input processors, and generates a range of actions that it sends to the appropriate (model-of-the-world) "actuators".

- **Inner speech** in a human is what the **mind's tongue** speaks and the **mind's ear** hears (when one talks in **inner voice** to oneself). **Inner speech** enables **CTM** to recollect its past, predict its future, and make plans.[41] The gists of **inner speech** (such as occur in talking to oneself or hearing in a dream) are nearly indistinguishable from the gists of **outer speech** (the gists created by the **Input maps**).[42]

---

[37] **Inner speech** is the inner voice that **CTM** uses to do planning and forecasting.

[38] Recall (Section **1.1**) that **inner speech** is always in Brainish and that **Input maps** turn **outer speech** into Brainish.

[39] Since blind people are conscious, a **Visual Scene processor** is not necessary for consciousness: A **Sensory Scene processor**, for example, can replace it.

[40] (Hurlburt & Heavey, 2015) identify five types of thoughts (gists) that humans are conscious of: an inner voice (an articulation of one's thoughts), an inner image (perhaps a map or dream image), a sensation (mostly external, as of hot, cold, tasty, slippery), a feeling (mostly internal, as of joy, anger, desire), a wordless thought (mindful meditation). As discussed in this chapter, the processors that produce such gists play a special role in giving **CTM** its sense of consciousness.

[41] Many animals have speech, not just humans**.** For example, prairie dogs have a complex communication system of tones for communicating information about predators to other prairie dogs (Slobodchikoff, Perla, & Verdolin, 2009) and (Slobodchikoff C. N., 2012). They use inner versions of this language to plan how to avoid predators.

[42] In humans, inner speech sounds so much like outer speech that it can be difficult, as in schizophrenia, to distinguish between Inner and outer speech (Rosen, et al., 2018).

- **CTM**'s **inner vision** enables **CTM** to create the **inner images** that **CTM** uses to generate imaginings or dreams. Examples of imaginings include maps, visual concepts, and so forth. The gists of **inner vision** are barely distinguishable from the gists of **outer vision** (the gists created by the **Input maps**).[43] Most humans see inner images most sharply in dreams, and with notable exceptions much less sharply in daydreams or imaginings (Marks, 1973) and (Zeman, Dewar, & Della Sala, 2015).[44] Whether awake or in a dream, an image gist can give the impression of an entire scene, all that the eye sees. After all, the gist holds the essence of the scene. The impression of seeing the whole of a scene is an illusion.

In summary, the *feeling* of consciousness or conscious awareness arises in part from the fact that Brainish is a very expressive language, and that every chunk that ever reaches **STM** is heard by **CTM**'s "inner ear", seen by its "inner eye", felt by its "inner sense of touch" and so on, in ways that very closely match what is heard, seen, and felt by outer ears, eyes, and touch.[45] This makes dreams extraordinarily realistic. Reduced functioning of these processors may create a reduced sense of consciousness. For example, Jill Bolte Taylor in My stroke of insight (Taylor, 2008) describes her own diminished state of consciousness following a stroke that disabled her (inner and outer) speech centers.

3. We argue that **CTM's** continuous cycling through prediction, feedback and learning (Section **1.5**), together with the **stream of consciousness** (Section **1.6**), play a role in **CTM**'s feeling of consciousness. This constant pro-active prediction-making and subsequent action informed by feedback that helps give this feeling. The feeling is further enhanced by (parallel) predictive dynamics in **CTM**'s **Model-of-the-World** where planning and testing is constantly carried out, often before action is taken by the **CTM**. Positive feedback gives **CTM** an indication that it understands what is going on; negative feedback - unless it is about something that could not have been predicted such as an unexpectedly loud noise - gives **CTM** evidence of something that it did not know or understand.
4. A minimal **general ability to think/plan** plus the **motivation (= energy + drive)** to do it. As (Valiant, 2013) says, "While there may be many kinds of intelligence, some minimum ability to reason from learned information, with all the uncertainties that that entails, has to have a role."

We now look at the **CTM** from the point of view of the **Model-of-the-World processor.** That processor like all processors is aware of both the **inner world** of imaginings and dreams (which it gets from **STM**) and the **outer world** (which it gets indirectly from **STM**), hardly distinguishing between outer and inner languages and sensations. Additionally, the **Model-of-the-World processor** incorporates and tags, as appropriate, this information in its various **model(s)-of-the-world**, including tagging the "**CTM"** in all its **models-of-the-world** as "consciously aware". From outside **CTM**, we see that something about **CTM** is conscious. It cannot be the **Model-of-the-World processor** itself or any other processor, as processors are just machines running algorithms. We propose that the view that **CTM** as a whole is conscious, as normally understood, is a consequence in part of the fact that the **Model-of-the-World processor** views the "**CTM**" in its **models-of-the-world** as conscious.

---

[43] To thwart schizophrenic hallucinations, the human brain needs to distinguish inner images from outer images when awake. The brain has various tricks for doing this, one being to make dreams hard to remember.

[44] The eminent architect, Tasso Katselos, has written us (personal communication) that "I am constantly generating images in my mind, [but] for me their final planning and execution never reach the richness and detail of the dream." On the other hand, in "An update on 'extreme imagination'", Adam Zeman says "that around 2-3% of the population, with *aphantasia*, lack a mind's eye, and that a somewhat larger percentage, with *hyperphantasia*, have imagery that is 'as vivid as real seeing'" (Zeman A. , 2020).

[45] The same can be said for other senses like (dolphin's) sonar, (bee's) ultraviolet vision, and (dog's) sense of smell.

# 4 The Hard Problem for Pain and Pleasure

"Nature has placed mankind under the governance of two sovereign masters, pain and pleasure." - Jeremy Bentham (1776)

David Chalmers (Chalmers, 1995) has defined the *Easy* and *Hard Problems* of consciousness which we reformulate here as follows:

- The **Easy Problem**: Make a robot that **simulates** feelings.
- The **Hard Problem**: Make a robot that truly **experiences** feelings.

While Chalmers is interested in all qualia,[46] we restrict our discussion of the hard problem to the qualia of pain and pleasure, including their extremes of agony and ecstasy. We do this in part to narrow the problem and in part because the generation of these particular feelings is especially mystifying. While the explanations for pain also work reasonably well for fear, the explanations for pleasure are necessarily different.

While current day robots may simulate emotions, they do not suffer the agony of pain, nor do they delight in joys and pleasures. We want an explanation for pain and pleasure that works as well for robots having brains of silicon and gold as for animals having brains of flesh and blood. [47]

The next two sections present our explanations for pain and pleasure: 4.1 deals with pain, **4.2** with pleasure. Although the model provides a measure of insight, we do not claim to have all the answers.

We start with **pain**. To clarify the difficulty of the hard problem for the case of pain, we describe a disorder called **Pain-Asymbolia**. (The corresponding disorder for the case of pleasure, called **Anhedonia**, is relevant to Section **4.2** below.) **Pain-Asymbolia** is a disorder in which the individual knows all there is to know of her pain, but she does not suffer from it (Bain, 2016). We distinguish two types:

**Pain-Asymbolia 1**. When hurt, the asymbolic person shows no outward sign of pain. She does not grimace or cry out under pain; she typically giggles when pinched and pricked.

**Pain-Asymbolia 2**. When hurt, the person shows outward signs of pain. She grimaces, cries out, etc. Notwithstanding these observable signs, the pain does not cause any suffering.

Both types of pain-asymbolics have working nociceptors (sensory receptors for painful stimuli). They are as aware of their pain as any normal human being: its location, its intensity, whether it is burning-hot or freezing-cold, and so on. But… they do not suffer.

Current-day robots are pain-asymbolic (and anhedonic).

---

[46] **Qualia = Individual instances of subjective, conscious experience.** For example, qualia include the concept of the **color** blue; the experience of **seeing** blue eyes; **hearing** "Rhapsody in Blue"; and **feeling** blue.

[47] Because it is possible for a machine to simulate pain and pleasure, we argue that any explanation for feelings in a machine needs to include knowledge of how the machine works. For comparison, consider Gordon (not George) Gallup's mirror test for self-awareness, which does a reasonably good job of testing for self-awareness in visually oriented animals (Gallup, 1977). The mirror test might be a good test for self-awareness in visually oriented machines as well, except that it is easy to build a machine with no self-awareness that passes the mirror test. For this reason, any test of self-awareness in a machine needs to include knowledge of how the machine works. Similarly, we claim that a test for deciding if a machine can feel/experience pain or pleasure needs to understand how the machine works.

## 4.1 Pain

Our primary reason for wanting to understand pain is to solve the puzzle of how nature produces the *feeling* of pain.[48] A related but different reason is to figure out how a **CTM** can experience the feeling of pain.

But why would we want a **CTM** or robot to experience the feeling of pain?

- For the same reason animals feel pain. Animals born without the suffering of pain don't live long. Children born with pain asymbolia typically live no more than 3 years.
- Because we want robots to have empathy. Humans find it hard to understand a feeling if they have never had that feeling.
- We also hope that this understanding of pain will enable humans either to control it in themselves or understand why they cannot.[49]

We have five suggestions for pain. Only five:

1. **Extreme pain** occurs when a chunk of extreme pain, a "scream of pain" in Brainish[50], takes over **STM**. Its great **intensity** makes it impossible for other chunks to compete successfully for **STM** - unless they too have comparably great **intensity**. Of equal importance, pain's weight, having a negative valence, makes each processor assign correspondingly great weight to reducing the pain.

   **When extreme pain messages are broadcast** from **STM**, every processor spends a fraction of time proportional to the intensity of the pain to find a way to ease it. At a minimum, processors are programmed to store information that relates the pain to their own capabilities. For example, the processor for recognizing faces can store whatever faces appear concurrently with the pain, and whether those faces are mitigating or exacerbating the pain. The processor for sleep can let the pain affect sleep by making it harder/easier to sleep. The processor for sex can link pain to sex, making sex nastier or more intense (a la Marquis de Sade).

   **Confirmation for extreme pain:**

   **a)** In describing physical pain, (Harnby, 2017) writes: "I've only been in agony a couple of times in my life, and I was good for nothing, rendered almost immobile. Reason left me. So did language."

   **b)** In *The Poppy Factory,* (Fairchild, 1987) describes an experience of extreme pain by: "suddenly you're

---

[48] The *feeling* of great hunger arises in part because a strong hunger signal allows nothing else to get into **STM**. **CTM** learns the *meaning* of hunger in part through its discovery that eating (filling its fuel tank) when hungry reduces the intensity of the hunger signal.

Feelings of pain, fear and hunger (but not pleasure, relief or satiety) are achieved in similar fashion. Consider hunger: the infant **CTM** has a fuel gauge processor to indicate how much fuel is in its system. The level of the fuel gauge determines the weight of its hunger gist. A weight below zero indicates hunger, and the lower (more negative) the weight, the greater the hunger.

[49] Here are some ways in which pain is special (Morrison, 2009):
- It's loud. Even small amounts shout over everything else.
- It's intrusive. You can't will it away. About the most you can do is match its intensity and drown it out.
- It makes you want to pull away. It's a flinch, abstracted. We suspect this is the "primitive op".
  All animal life flinches, even stuff too simple to have a brain.

To this we add:
- It's motivating.

[50] The "scream of pain" in Brainish contains much more information than is conveyed in normal English. Artists, novelists, and poets try their best to capture it. Louise Harnby (Harnby, 2017) points to William Fairchild's (Fairchild, 1987) *The Poppy Factory* for his description of extreme pain: "Then suddenly you're down and all movement stops like a jammed cine film. You're still screaming but now it's different. It's because of the pain and when you try to get up, your legs won't move. You don't know where you are. All you know is that you're alone and probably going to die. When you stop screaming and look up, the sky is dark and you can't hear the guns any more, only the sound of someone moaning softly. It takes a few moments before you realize it's yourself."

down and all movement stops like a jammed cine film. You're still screaming but now it's different. It's because of the pain and when you try to get up, your legs won't move. You don't know where you are. All you know is that you're alone and probably going to die. When you stop screaming and look up, the sky is dark and you can't hear the guns any more, only the sound of someone moaning softly. It takes a few moments before you realize it's yourself."

2. **Sudden extreme pain** - a ligament at the moment it is torn – **interrupts** all unconscious processors. The shock of pain is instantaneous. How does this excruciating pain come about?

   **a)** When a chunk reaches **STM**, it gets broadcast. If that chunk has **intensity ≥ ι**, where **ι** is the **interrupt constant** of **CTM** (Section **1.4.1**), the ensuing broadcast causes all **LTM** processors to put their current work on a stack to pay "maximum attention" to the cause of the interrupt. The sudden interruption of all processing systems - as from an unexpected whack on the head - registers as shock.

   **Confirmation for sudden extreme pain:**

   People are known to remember exactly where they were when they tore a ligament. The decision to store was not made consciously. That's what autobiographical memory does when it gets interrupted (Fivush, 2011).

   **b)** The difference between broadcasts and interrupts is this: A broadcast is an input to all processors that is handled by every processor each in its own way, including possibly by disregarding it. Interrupts, however, compel processors to put their work on a stack and pay (maximum) attention, as mentioned above, to the cause of the interrupt.

   **c)** In sudden extreme pain, nothing else enters **STM**. All consciousness is on pain. Nothing can get into **STM** to temper that pain.

3. **Less extreme pain** and **chronic pain** do not so much prevent other chunks from reaching the stage as make it "difficult" for them to reach it. In the **deterministic CTM**, the **difficulty** for a chunk to get into **STM** is measured by how much greater the chunk's **intensity** would have to be for it to get into **STM**. In the **probabilistic CTM**, the **difficulty** is measured by how much greater the chunk's **intensity** would have to be to get allotted a "suitably larger" share of time in **STM**.

   **Confirmation for chronic pain:**

   **a)** **In** "The Impact of Persistent Pain on Working Memory and Learning", (Ayres & Smith, 2014) write: "Participants that identified as experiencing pain for 6 or more months demonstrated clinically low levels of pain, but nevertheless performed significantly worse than pain-free participants on retention and transfer tests."

   **b)** In 2008, about 100 million people were affected by chronic pain costing the U.S. $560 to $635 billion (in 2010 dollars) combing health care costs, work missed and lower wages. (Gaskin & Richard, 2012)

4. **Fear causes and enhances pain**. "Fear of pain and coping strategies… are known to play an important role in the development and maintenance of pain." (Mittinty, et al., 2017)

5. **Vicious cycles**. Concentration on pain reduces pain, permitting fear to take its place; concentration on fear reduces fear, permitting pain to take its place. Such a vicious cycle sustains and reinforces both pain and fear.

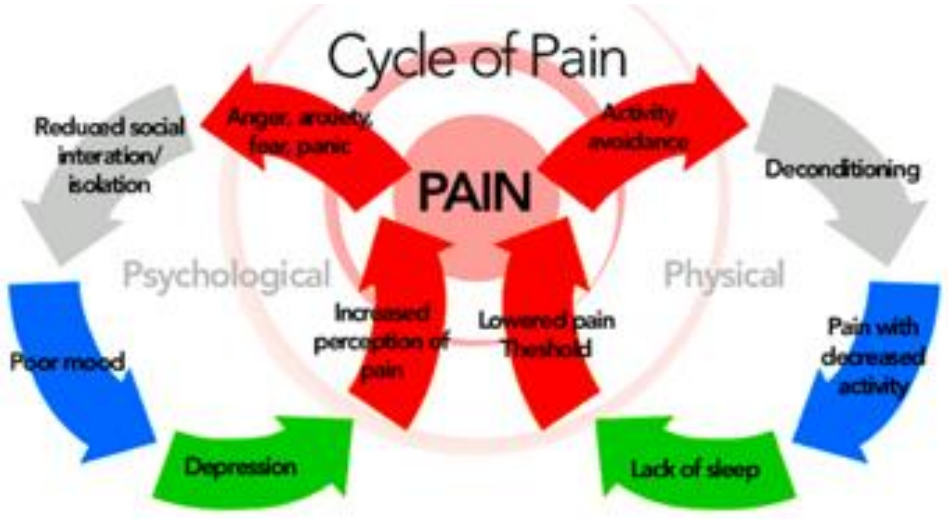

**Figure 7 Cycle of Pain. (Reproduced with permission from Co-Kinetic.com).**

## 4.2 Pleasure

Pain and pleasure are often viewed as opposite sides of the same coin:

> Opposite sides: "Losses are felt as pain or anxiety, and gains as pleasure." (Szasz, 1957)
>
> Same coin: "Emerging evidence from pain and reward research points to extensive similarities in the anatomical substrates of painful and pleasant sensations." (Leknes & Tracey, 2008).

Examples of pleasure include: • a mother's love; • avoidance of pain;[51] • success in achieving a goal - any goal that **CTM** has consciously set for itself;[52] • coming up with a new idea for a promising course of action.[53]

A processor expresses pleasure by doing some of what a processor for pain does when it gets a chunk on stage: it hogs the stage. This is particularly true when the pleasure is extreme (ecstasy).

There are also major differences between pain and pleasure.

In the **CTM**, every processor spends a fraction of its time looking for ways to *increase* its pleasure and *decrease* its pain (symbolized respectively by positively and negatively weighted gists). See Section **1.4.2**. Much of this is built into all **LTM** processors at birth, and reinforced by learning mechanisms. A child learns from its built-in suckle response that milk reduces pain – in this case the pain of hunger. The child generalizes the power of food to reduce the pain of hunger to its power to reduce all pains, and consequently it suckles whenever it has pain "even when the pain has nothing to do with hunger per se" (Leknes & Tracey, 2008).[54]

Similarly, having learned that positive moods counter negative moods, **CTM** can try to lessen any negative mood (pain, hunger, etc.) by seeking a counter-balancing positive mood (like food). In this way **CTM** *learns* that negative moods can be countered by positive moods. Unless a child is taught to avoid pleasure, she will generally not seek pain to counter-balance pleasure. This is one way in which pleasure and pain are not symmetrical. These and other (asymmetrical) functionalities account for why the feelings generated by positively and negatively weighted gists are different.

---

[51] For Epicurus, "happiness was the complete absence of bodily and especially mental pains" (Bergsma, Poot, & Liefbroer, 2008). But sometimes, pain can facilitate pleasure "by providing an important contrast for pleasurable experiences, increasing sensitivity to sensory input, and facilitating self-rewarding behavior" (Bastian, Jetten, Hornsey, & Leknes, 2014).

[52] "Nothing succeeds like success" ["Rien ne réussit comme le succès"] (Dumas, 1854); and "Success breeds success" (van de Rijt, Kang, Restivo, & Patil, 2014).

[53] Nevertheless, an excess of ideas - too many ideas to think through carefully - may be a sign of **mania,** while a dearth of ideas -too few to maintain interest- may be a sign of **depression**.

[54] In general, "the pleasure system relies on the *balanced interaction* over time of key brain regions…" Additionally, in humans, there appears to be a **"** 'common currency' reward network of interacting brain regions. Pleasures of food, sex, addictive drugs, friends and loved ones, music, art, and even sustained states of happiness can produce strikingly similar patterns of brain activity**"** (Kringelbach & Berridge, 2017).

As Michel Cabanac points out (Cabanac, 2013), behavior in humans and animals is "motivated by the trend to seek pleasure and avoid displeasure." Behavior is also "adapted to the defense of homeostasis in the long term, …, not limited to correcting immediate needs but also anticipated future needs for chemical and thermal energy."

Similarly (Lewis & Cañamero, 2016), studying autonomous robots, suggest that pleasure can have a "purely hedonic quality not directly linked to need satisfaction," and that both [hedonic and need] have "value for homeostatic management in terms of improved viability, as well as in terms of more flexibility in adaptive behavior."

We hypothesize that a system such as **CTM**, built to attain homeostasis via its inherent predictive dynamics (Section**1.5**), will have a feeling of pleasure and well-being in the quest and attainment of that goal.[55]

# 5 Summary

This paper looks at consciousness from the perspective of theoretical computer science. It presents a simple formal model of Bernard Baars' **GWT** (Global Workspace Theory) of Consciousness (Baars B. J., 1988) and (Baars B. J., 2019). We believe **GWT** captures the essence of consciousness. In formalizing our model, the **CTM** (Conscious Turing Machine), we also take a cue from (Turing, 1937), aiming for simplicity rather than complexity, for a simple model of consciousness rather than a complex model of the brain.

Even supposing we had a complete description of the brain, and the technology to duplicate it, that would not imply that we understand what gives rise to consciousness, the feelings of pain and pleasure, etc. We claim that what gives rise to consciousness is the expressiveness of the brain's inner language and the architecture of the system, basic processes and dynamics. One purpose of the **CTM** model, besides formalizing Baars' Global Workspace Theory, is to argue this claim. Another is to provide a theoretical computer science foundation for understanding consciousness.

This paper does the following:

1. It presents a simple mathematical model (Chapter **1**) of Bernard Baars' Global Workspace Theater of Consciousness – expressed formally in the definition of the **Conscious Turing Machine** (**CTM**).
2. It discusses **Brainish** (Section **1.1**), an enormously expressive language capable of generating the illusion of all sensations, actions, and feelings.
3. It defines a **chunk** (Section **1.3.1** and **1.3.2**) explicitly as a 6-tuple **<address, time, gist, weight, intensity, mood>**. Cognitive psychology generally identifies a chunk with the gist alone; the **CTM** chunk has an additional collection of numbers (address, time, weight, intensity, mood) that the human (only) vaguely or approximately senses.[56]
4. It defines the **conscious content** of **CTM** (Section **1.6**) to be whatever chunk is in **STM** (Short Term Memory); and then defines **conscious awareness** by **CTM** to be the reception by all **LTM** (Long Term Memory) processors of **STM**'s broadcast of that content. The **gists** of those broadcasts are the inner thoughts generated by **CTM**'s unconscious processors or the speech, vision, touch, taste, and/or whatever else is received as input by **CTM**.

---

[55] This aligns also with the hypothesis that **"**optimal metastability could be linked to a state of eudaimonia…. As such this objective measure of metastability could link eudaimonia (and emotions) to more global theories of brain function. One such example is the global neuronal workspace model ….**"** (Kringelbach & Berridge, 2017).

[56] See e.g., "Chunking Mechanisms and Learning" (Gobet & Lane, 2012).

5. It discusses how prediction, feedback and learning (Section **1.5**) enables **LTM** processors to adjust the weights they give their gists based on information gleaned from conscious and unconscious awareness of mistakes, inconsistencies[57], unexpected effects, and so on.

6. It explains why the continuous broadcasting of the content of **STM**, the **stream of consciousness**, helps give rise to consciousness as we know it (Chapter **3**), namely:
   - The feeling of consciousness arises in **CTM** because all its processors, including *especially those that are particularly responsible for consciousness*, are privy to the same (conscious) content of **STM**, and
   - the gists of outer speech (what we say and hear in the world), outer vision (what we see in the world), and so on, are nearly indistinguishable from the gists of inner speech (what we say to ourselves), inner vision (what we see in dreams), and most importantly,
   - the **gists** are expressed in Brainish, an enormously expressive language capable of generating the illusion of sensations, actions, and feelings.

7. It gives some understanding of how a pain or pleasure *experience*, not just its *simulation*, is produced (Chapter **4**).

In summary, we argue that the *feeling* of consciousness in the **CTM** is produced using the *expressive power of Brainish*, which is **CTM**'s inner language for describing all elements of both its outer and inner worlds, and by its *architecture,* certain *special processors*, and its *predictive dynamics* (prediction, feedback and learning).

An expanded version of this paper (Blum, Blum, & Blum, Towards a Conscious AI: A Computer Architecture Inspired by Cognitive Neuroscience, In preparation) will cover the topics presented here in considerably more detail, especially the **Sleeping Experts Algorithms**, as well as additional topics such as dreams, illusions and free will.[58]

# 6 Relation to Other Theories of Consciousness

The **CTM** is influenced by Baars' **GWT**, which is supported by the Global Neuronal Workspace Theory (**GNWT**) of (Dehaene & Changeux, 2011) and (Dehaene S. , 2014) in their investigation of neural correlates of consciousness. Like the **LIDA** model of cognition (Baars & Franklin, 2007) and (Baars & Franklin, 2009), **CTM** is architectural. Unlike **LIDA**, which is a more elaborate model of **GWT**, the **CTM** is intended to be a minimal model of **GWT** sufficient to explain a wide range of conscious phenomena.

We see a kinship between the **CTM** and the self-aware robots developed by (Chella, Pipitone, Morin, & Racy, 2020).

Philosophically, we align with much of Daniel Dennett's functionalist perspective (Dennett D. C., 1991)[59] and not so much with David Chalmers' phenomenalist focus on **qualia** (Chalmers, 1996). Along with Dennett, we do not see the *explanatory gap* (Levine, 1983) as insurmountable. Indeed, we see the **CTM** as helping to explain the feeling of "what it is like" (Nagel, 1974).

[57] Inconsistencies are detected by the **Model-of-the-World** processors (Chapter **3** item 2), among others.

[58] Topics will also include: a further discussion of **LTM** processors (some central and some not so central for consciousness) including **Sleep** and **Dream Creation** processors, **Motivation** processors, **Meditation** processors; processor recruitment; evolution of processors; more explanatory examples; how **CTM** understands and extends its understanding in new directions; how various functionalities (like the function of a Central Executive) emerge without being built into the model; and a further discussion of computational and complexity measures (numbers, size and time).

[59] We generally agree with Dennett except for his view that we are the only species to have consciousness, see (Dennett D. C., 1978), and more recently (Dennett D. C., 2018). In an otherwise excellent interview with Louis Godbout, Dennett expresses his view that a dog does not have conscious awareness since it "can't tell a story about what it is thinking about; it can't, it doesn't have language" (Dennett D. C., 2019).

As in Michael Graziano's Attention Schema Theory (**AST**) (Graziano, Guterstam, Bio, & Wilterson, 2020), **CTM** is consciously aware of both external and internal events. Both **AST** and **CTM** appear to embody and substantiate **illusionist** notions of consciousness proposed by Dennett (Dennett D. C., 2019) and Keith Frankish (Frankish, 2016).[60] Basic **AST** is similar to **GWT**: its i-consciousness (i for information) aligns somewhat with **CTM**'s conscious awareness.[61] We do not agree with Graziano et. al. that **GWT** "leaves unexplained how people end up believing they have subjective experience," i.e. leaves an explanatory gap. In imaginings and dreams, for example, the feeling of subjective experience in the **CTM** arises when the "winning chunks" of those imaginings and dreams are received by the same (unconscious) processors that receive chunks directly from the environment via **Input maps**. Additionally, the **Model-of-the-World processor** incorporates the information gotten from the winning chunks (i.e., the **conscious content** of the **CTM**) into its **model(s)-of-the-world**, as appropriate, tagging the "**CTM"** in all **models-of-the-world** as "consciously aware". Such experiences get additional heft from the constant bubbling of chunks into **STM**, and their broadcast to **LTM**, forming streams of consciouness (Section **1.6** and Chapter **3**). These chunks are constantly evolving due in part to **CTM**'s dynamics of prediction, feedback, and learning (Section **1.5**).

With respect to its predictive dynamics (Section **1.5**), **CTM** incorporates elements similar to the "predictive processing" (**PP**) of (Lee & Mumford, 2003), (Friston, 2003), (Cleeremans, 2014), (Clark, 2015), (Seth, 2015) and (Hohwy & Seth, 2020). See also Section **4.2** on pleasure.

By utilizing existing technology (or apps) to supplement its supply of **LTM** processors (see Section **1.7**), **CTM** incorporates elements similar to those advocated by (Clark & Chalmers, 1998)'s "extended minds".

We agree with Christof Koch that "There isn't a Turing Test for consciousness. You have to look at the way the system is built. You have to look at the circuitry, not [only] its behavior" (Paulson, 2017). We would emphasize "architecture" as well as "circuitry".

Along these lines, Integrated Information Theory (**IIT**), the theory of consciousness developed by Giulio Tononi, (Tononi, 2004) and supported by Koch (Tononi & Koch, 2015), proposes a measure of consciousness called **PHI**, defined using Shannon's information theory. Tononi proposes five "axioms" (properties) necessary for any causal system to have consciousness.[62] Given a detailed specification of a **CTM**, one could in principle compute its **PHI** and compare it to the **PHI** of any other precisely defined causal system. It turns out that many causal physical systems have non-zero measures of **PHI**.[63, 64]

---

As for animal consciousness, we agree with (Mumford, submitted 2019) that consciousness is a matter of degree. Here he cites (Merker, 2007) that consciousness does not need a cerebral cortex, it arises from midbrain structures. We would also cite other studies, e.g., (Slobodchikoff C. N., 2012).

[60] Saying that the feeling of consciousness is an illusion does not deny the existence of that feeling. As a familiar example, the fact that a movie is made up of (many) discrete still images does not affect the feeling of continuity one gets from viewing it. The feeling of continuity is an illusion.

[61] Full **AST** has three neural networks (A for receiving information, B for constructing an attention schema, and C for reporting to the outside world) to obtain a system which purportedly thinks it has subjective experience (m-consciousness, m for mysterious).

[62] In (Koch, 2019), Christof Koch outlines the axioms: "[E]very conscious experience has five distinct and undeniable properties: each one exists for itself, is structured, informative, integrated and definite".

[63] **IIT** in particular validates animal consciousness.

[64] There are a number of "popular" books relating to consciousness. A few that we have particularly enjoyed include *A Leg to Stand On* (Sacks, 1984, 1993); *Moonwalking With Einstein: The Art and Science of Remembering Everything* (Foer, 2011); *The Man who wasn't there* (Ananthaswamy, 2015); *Other Minds* (Godfrey-Smith, 2016); and *Mamma's Last Hug* (de Waal, 2019).

## Acknowledgements

We are especially grateful to Jean-Louis Villecroze for his comments, suggestions, and painstaking multiple reviews of this paper, his pointers to the literature, and his ongoing work to simulate **CTM** (Villecroze J. L.), (Villecroze J.-L. , 2019). We thank Paul Liang for his continuing work with us. We thank the students and faculty at CMU and PKU for their feedback in our courses. We are grateful to our friend Michael Xuan for his enormous personal support and encouragement. We thank UniDT for their supporting grant of our work.

## **About the Authors** of the expanded version of this paper **(Blum, Blum, & Blum, In preparation)**

**Manuel** has been motivated to understand the mind/body problem since he was in second grade when his teacher told his mom she should not expect him to get past high school. As an undergrad at MIT, he spent a year studying Freud and then apprenticed himself to the great anti-Freud[65] neurophysiologist Warren S. McCulloch, who became his intellectual mentor. When he told Warren (McCulloch) and Walter (Pitts) that he wanted to study consciousness, he was told in no uncertain terms that he was verboten to do so - and why. As a graduate student, he asked and got Marvin Minsky to be his thesis advisor. Manuel is one of the founders of complexity theory, a Turing Award winner, and has mentored many in the field who have chartered new directions ranging from computational learning, cryptography, zero knowledge, interactive proofs, proof checkers, and human computation. **Manuel Blum** mblum@cs.cmu.edu

**Lenore** has been passionate about mathematics since she was 10. She attributes that to having dropped out of school when she was 9 to wander the world, then hit the ground running when she returned and became fascinated with the Euclidean Algorithm. Her interests turned to non-standard models of mathematics, and of computation. As a graduate student at MIT, she showed how to use saturated model theory to get new results in differential algebra. Later, with Mike Shub and Steve Smale, she developed a foundational theory for computing and complexity over continuous domains such as the real or complex numbers. The theory generalizes the Turing-based theory (for discrete domains) and has been fundamental for computational mathematics. Lenore is internationally known for her work in increasing the participation of girls and women in STEM and is proud that CMU has gender equity in its undergraduate CS program. **Lenore Blum** lblum@cs.cmu.edu

**Avrim** had an earlier start than the elder Blums. He spent his first two years at MIT, in his mom's office in the Math Department, and in his dad's office in McCulloch's lab. In sixth grade, he solved an extra credit math problem by programming his home-made computer to get a feel for the problem, then (once he saw what was going on) stated and proved the desired result. Because he used a computer, he got no credit. Odd, because he was pointing to a novel way (at the time) to solve a math problem. Avrim's expertise is Machine Learning Theory. He has been an advisor to many of the young leaders in the field. **Avrim Blum** avrim.blum@gmail.com

All three Blums received their PhDs at MIT and spent a cumulative 65 wonderful years on the faculty of the Computer Science Department at CMU. Currently the elder two are emeriti and the younger is Chief Academic Officer at TTI-Chicago, a PhD-granting computer science research institute focusing on areas of machine learning, algorithms, AI (robotics, natural language, speech, and vision), data science and computational biology, and located on the University of Chicago campus.

This is their first joint paper.

---

# Addendum: High Level Explanations

In this paper we explored explanations for the feelings of pain and pleasure in the **CTM** (Chapter **4**). In *Insights from the Conscious Turing Machine* (Blum & Blum, 2021) we consider additional phenomena generally associated with consciousness (see https://arxiv.org/pdf/2107.13704.pdf). These include examples related to vision (blindsight, inattentional blindness, and change blindness) as well as dreams, free will and altered states. We give explanations *derived from the formal model* and draw confirmation from consistencies *at a high level* with the psychological and neuroscience literature. The model and explanations are developed further in (Blum, Blum, & Blum, In preparation).

[65] Where Freud had written *The Future of an Illusion* (Freud S. , 1927), McCulloch followed with "The Past of a Delusion" (McCulloch W. S., 1953).